\documentclass[lettersize,journal]{IEEEtran}
\usepackage{amsmath,amsfonts}
\usepackage{amssymb, bm}
\usepackage[ruled,vlined]{algorithm2e}
\usepackage{algorithmic}
\usepackage{array}
\usepackage[caption=false,font=normalsize,labelfont=sf,textfont=sf]{subfig}
\usepackage{textcomp}
\usepackage{stfloats}
\usepackage{url}
\usepackage{verbatim}
\usepackage{graphicx}
\usepackage{cite}
\usepackage{multirow}
\usepackage{multicol}
\usepackage{makecell}
\usepackage{booktabs}
\usepackage{supertabular}

\usepackage{pdflscape} % For landscape mode
\usepackage{array} % For advanced column formatting

\begin{document}

\title{Accelerated Training of Federated Learning via Second-Order Methods}

\author{Mrinmay Sen, Sidhant R Nair,  C Krishna Mohan,~\IEEEmembership{Senior Member,~IEEE}
\thanks{Mrinmay Sen is with the Department of Artificial Intelligence, Indian Institute of Technology Hyderabad, India and the Department of Computing Technologies, Swinburne University of Technology, Hawthorn, australia (e-mail: ai20resch11001@iith.ac.in).}
\thanks{Sidhant R Nair is with the Department of Mechanical Engineering , Indian Institute of Technology Delhi, India.}
\thanks{C Krishna Mohan is with the Department of Computer Science and Engineering, Indian Institute of Technology Hyderabad, India.}

\thanks{Manuscript received April 19, 2021; revised August 16, 2021.}}

\markboth{Journal of \LaTeX\ Class Files,~Vol.~14, No.~8, August~2021}%
{Shell \MakeLowercase{\textit{et al.}}: A Sample Article Using IEEEtran.cls for IEEE Journals}


\maketitle
\begin{abstract}
This paper explores second-order optimization methods in Federated Learning (FL), addressing the critical challenges of slow convergence and the excessive communication rounds required to achieve optimal performance from the global model. While existing surveys in FL primarily focus on challenges related to statistical and device label heterogeneity, as well as privacy and security concerns in first-order FL methods, less attention has been given to the issue of slow model training. This slow training often leads to the need for excessive communication rounds or increased communication costs, particularly when data across clients are highly heterogeneous. In this paper, we examine various FL methods that leverage second-order optimization to accelerate the training process. We provide a comprehensive categorization of state-of-the-art second-order FL methods and compare their performance based on convergence speed, computational cost, memory usage, transmission overhead, and generalization of the global model. Our findings show the potential of incorporating Hessian curvature through second-order optimization into FL and highlight key challenges, such as the efficient utilization of Hessian and its inverse in FL. This work lays the groundwork for future research aimed at developing scalable and efficient federated optimization methods for improving the training of the global model in FL.

\end{abstract}
\section{Introduction}
\IEEEPARstart{F}{ederated} Learning (FL), first proposed by \cite{mcmahan2016federated} (namely FedAvg), represents a transformative approach to distributed machine learning, where data privacy and locality constraints render traditional centralized methods infeasible. In FL, clients collaboratively train a shared global model while keeping their data local, which creates unique optimization challenges due to the decentralized and often heterogeneous nature of the data. In FL, each client trains the shared global model using its local data and local optimizer. The host server then collects all the locally trained models and aggregates them to update the global model, which is subsequently sent back to all the clients. This iterative process continues for $T$ communication rounds until the desired performance is achieved by the global model. Due to the computation of only gradients, FedAvg incurs a per-sample local computation cost of $\mathcal{O}(d)$ and a memory cost of $\mathcal{O}(2d)$ for storing both the local model and gradient. For aggregating local models, FedAvg incurs computation costs of $\mathcal{O}(\mathcal{K}d)$ in the server. The memory requirements for the server in FedAvg are $\mathcal{O}(\mathcal{K}d)$ for storing all the local models during aggregation. As FedAvg requires sharing only model parameters from clients to the server and from the server to clients, it has a per communication transmission cost of $\mathcal{O}(2d)$. Here, $d$ is the number of model parameters and $\mathcal{K}$ is the number of clients. Even though FedAvg is efficient in terms of computation $\&$ memory costs as well as transmission costs, there are some major challenges with FedAvg. The practical application of FedAvg faces several major challenges, including data heterogeneity across clients \cite{MathewAfedprox_24,KarimireddySCAFFOLD_20,li2021_MOON,AcarZNMWS21feddyn,barba2021implicitFedGA,nagaraju2023handling}, partial client participation in each communication round due to device label heterogeneity \cite{yang2023littleSAFL,yang2021achievinglienar_speed,gu2021fast,wu2023anchor,jhunjhunwala2022fedvarp}, differential privacy \cite{wei2020federated_diff1,truex2020ldp,adnan2022federated,triastcyn2019federated,mohammadi2021differential}, adversarial attacks \cite{wang2020attack,darzi2024exploring,song2020analyzing,hu2023toward,chen2020zero}, and communication overhead caused by an increased number of communication rounds due to slow training of the global model \cite{WangNeurIPSgiant_18,guptalocalnewton_21,Mafedsso_22,liDANE_19,safaryanfednl_22,mrinmaynys-fl_23,nagarajufonn_23,yangover_the_air_22,senfoplahd_23,dinhdone_22,senfreng_23}. Mathematically, FL can be framed as minimizing a global objective function:
\begin{equation}
    \min_{\textbf{x}} F(\textbf{x}) = \frac{1}{\mathcal{K}} \sum_{k \in \mathcal{K}} f_k(\textbf{x})
    \label{Eq:1}
\end{equation}
where \( \mathcal{K} \) denotes the set of participating clients, \( f_k \) represents the local loss function for client \( k \), and \( \textbf{x} \in \mathbb{R}^d \) is the global model shared across all clients. The solutions to Problem \ref{Eq:1} can be categorized into two groups based on the order of the Taylor series approximation of the objective function: the first group comprises first-order FL methods and the second group comprises second-order FL methods. First-order FL methods suffer from slow convergence because they rely solely on the gradient of the global objective during model updates. Although these methods are computationally and memory efficient, as they only require gradient computation, their slower convergence can hinder the overall training process. In contrast, second-order FL methods enhance convergence by utilizing both the gradient and the Hessian matrix during global model updates. This dual approach results in quadratic convergence and reduces the number of communication rounds required in FL. Despite the advantages of second-order optimization in improving global model convergence, challenges related to the Hessian matrix and its inverse remain. These challenges motivate researchers to develop efficient algorithms that make effective use of the Hessian during global model updates. Efficient utilization of the Hessian and its inverse is crucial for minimizing communication overhead, particularly in the context of second-order optimization in FL.

In this paper, we investigate the challenge of slow training of the global model caused by the use of first-order optimization methods. To address this issue, we explore various second-order FL methods designed to accelerate global model training. While numerous surveys have examined FL, most have focused on challenges such as heterogeneous data across clients, partial client participation, differential privacy and adversarial attacks, typically in the context of first-order methods with linear convergence rates. However, less attention has been given to the issue of increased communication rounds—and the corresponding higher communication overhead—resulting from the slow convergence of first-order methods. In this survey, we aim to evaluate the effectiveness of various second-order FL algorithms in mitigating slow convergence of the global model, with the indirect goal of reducing communication rounds. We categorize these methods based on their approach to Hessian utilization, including Hessian-free approach, Full Local Hessian Computation, Quasi-Newton based, Hessian diagonal approximation, Nyström method-based Hessian approximation and One-rank approximation of Hessian. These methods are then compared based on key factors such as computational and memory costs, transmission costs, convergence rates and their respective advantages and disadvantages. Additionally, we provide empirical analyses to support our comparisons. With this survey, researchers can make informed decisions about which algorithms to apply under specific conditions and gain valuable insights into potential areas for further research to develop more efficient second-order FL methods, addressing the limitations of existing approaches. 

\subsection{Prior Surveys on FL}
While numerous surveys have examined various aspects of FL, from privacy mechanisms to communication protocols, there remains a notable gap in the comprehensive analysis of second-order optimization methods in FL for improving convergence. This gap is especially significant, given recent breakthrough developments in accelerated training techniques that could fundamentally reshape how we approach federated optimization. \cite{Banabilahsurvey2_22}, \cite{kumar2023impact} \cite{zhang2023systematic}, \cite{shan2024differential}, \cite{fu2024differentially}, \cite{feng2025survey} and \cite{akhterchallengessurvey} establish comprehensive frameworks for understanding FL architectures, privacy mechanisms and design aspects. However, their treatment of optimization techniques primarily focuses on first-order methods, leaving advanced optimization strategies as a secondary consideration. A series of focused surveys by \cite{zhusurvey4_21}, \cite{lufederatedsurvey_24}, \cite{hansurvey6_22}, \cite{HuangYSWLDY24FL_survey} and \cite{zhaofederatedsurvey_18} have extensively examined the non-IID data challenge in FL. These works primarily concentrate on characterizing and addressing data heterogeneity through statistical approaches and client selection strategies, rather than exploring optimization-based solutions. While they thoroughly analyze how non-IID data affects model convergence, they largely overlook the potential of second-order optimization methods in accelerating FL training under heterogeneous conditions. Their emphasis on data-centric solutions, while valuable, leaves unexplored the promising intersection of advanced optimization techniques. This gap is particularly significant given that second-order methods could potentially provide more efficient solutions by leveraging curvature information to better navigate the heterogeneous loss landscape.
Key theoretical works like \cite{bischoffsurvey_21} provide valuable insights into second-order optimization in FL. However, their pre-2022 publication dates mean they precede crucial methodological breakthroughs. This temporal gap highlights the need for analysis incorporating these recent advances. \cite{tan2019review} and \cite{kashyap2022survey} investigate different second-order techniques in the context of centralized learning. 
\cite{Ma_stateoftheartsurvey3_22} has illuminated FL applications across various domains. While effectively categorizing data partitioning approaches, this work doesn't address how emerging second-order methods could transform domain-specific implementations.
The existing survey landscape reveals three critical gaps:\\
$\bullet$ Insufficient coverage of recent second-order optimization methods in FL and their theoretical foundations\\
$\bullet$ Lack of unified taxonomies for categorizing and comparing different second-order approaches\\
$\bullet$ Limited systematic benchmarking of computational, communication, and convergence trade-offs
\subsection{Contributions}
In this paper, we introduce a detailed survey on second-order optimized federated learning methods, focusing on their theoretical foundations, algorithmic innovations, and empirical performance. Our primary contributions are as follows:\\
\textbf{Unified Categorization:}
We systematically group and analyze second-order optimization techniques, categorizing them based on their underlying approaches such as diagonal approximations, low-rank methods, Hessian-free techniques, preconditioning, second-order expansions, and more.\\
\textbf{Comprehensive Benchmarking:}
We provide a detailed evaluation of these methods, emphasizing key metrics such as computation $\&$ memory requirements, transmission cost, convergence properties, mathematical insights, and Hessian approximation method in federated learning settings.\\
\textbf{Future Research Directions:}
We identify open challenges in second-order federated optimization, including scalability to larger datasets, balancing computational and communication costs, and improving robustness in heterogeneous environments.

\section{Background}

In this section, we describe optimization techniques based on Taylor series approximations of the objective function and their limitations in FL.

\begin{figure}[ht]
    \centering
    \begin{minipage}{0.45\textwidth}
        \centering
        \includegraphics[width=\linewidth]{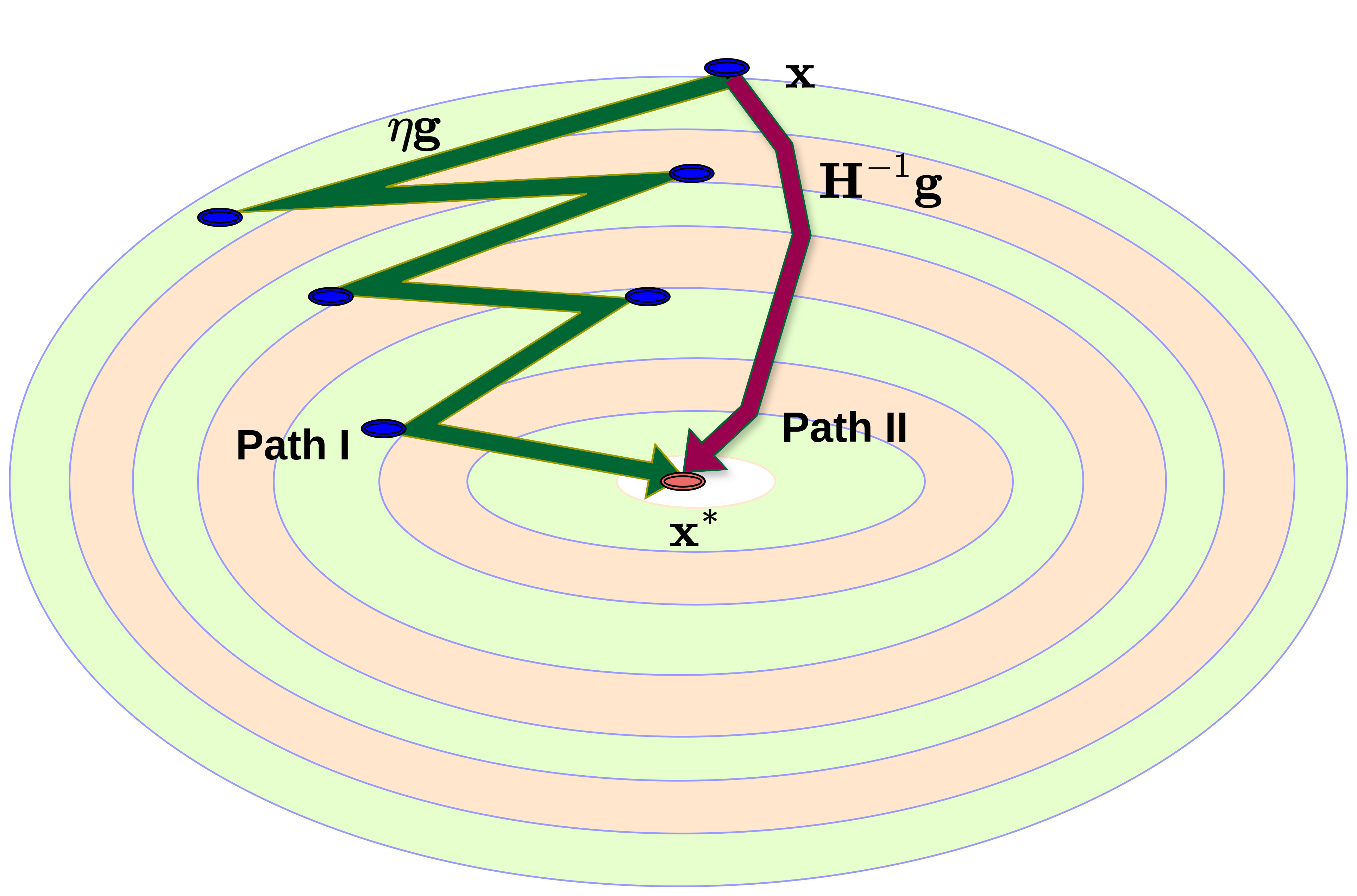}
    \end{minipage}
    \caption{This figure compares first-order and second-order optimization methods. The green path (Path I) represents the trajectory taken by the first-order optimization technique to reach the optimum, while the red curve (Path II) illustrates the second-order optimization technique, which directly converges to the optimum by following the curvature.}
    \label{fig:fig1}
\end{figure}

\subsection{Taylor Series Approximation Based Optimization Techniques}
The optimization methods for solving Eq. \ref{Eq:1} can be categorized into two groups based on the order of the Taylor series approximation of the objective function $F(\textbf{x})$, namely first-order optimization and second-order optimization (Newton method of optimization). At any point $\textbf{x}$, the Taylor series can be formulated as follows :

\begin{equation}
    F(\textbf{x} + \Delta \textbf{x}) \approx F(\textbf{x}) +  {\Delta \textbf{x}}^T \nabla F(\textbf{x}) + \frac{1}{2} {\Delta \textbf{x}}^T \nabla^2 F(\textbf{x}) {\Delta \textbf{x}} + ....
    \label{Eq:2}
\end{equation}
\subsubsection{First-Order Optimization}
From Eq. \ref{Eq:2}, at point $\textbf{x}$, first-order optimization is derived as follows, where the model update step is calculated by scaling the gradient ($\textbf{g} = \nabla F(\textbf{x})$) of the objective function computed at this point with a fixed step size, i.e., learning rate ($\eta$), as shown in Eq. \ref{Eq:3} \cite{ketkarSGD_17}. Here, the point refers to the set of model parameters $\textbf{x} \in \mathbb{R}^d$.

\begin{equation}
    \textbf{x} \leftarrow \textbf{x} - \eta \textbf{g}
    \label{Eq:3}
\end{equation}

This iterative process continues until we reach the optimum of the objective, i.e., $\textbf{x}^*$. Due to the direct scaling of the gradient with a fixed step size, the path to the optimum from the initial point $\textbf{x}$ becomes zigzag, and it requires an increased number of iterations to reach the optimum (see Path I in Figure \ref{fig:fig1}) \cite{kashyap2022survey}. The implementation of this first-order method in FL leads to an increased number of communication rounds while aiming for a targeted performance from the global model.

\subsubsection{Second-Order Optimization (Newton method of optimization)}
In contrast, second-order optimization (Newton method of optimization) is derived from the second-order Taylor series approximation, where the gradient is adaptively scaled by scaling it with the inverse of the Hessian ($\textbf{H} = \nabla^2 F(\textbf{x})$), as shown in Eq. \ref{Eq:4}. This approach helps follow the curvature of the objective function while searching for the optimum (see Path II in Figure \ref{fig:fig1}), leading to quadratic convergence and a reduced number of iterations. This quadratic convergence nature motivates researchers to replace first-order method with second-order method in FL to improve the convergence of the global model, thereby helping to reduce the number of communication rounds and communication overhead in FL.

\begin{equation}
    \textbf{x} \leftarrow \textbf{x} - \textbf{H}^{-1} \textbf{g}
    \label{Eq:4}
\end{equation}

\subsubsection{Challenges with Second-order Optimization in FL}

Even second-order optimization is beneficial in terms of improving convergence of the global model in FL, it has some drawbacks in computation of Hessian and its inverse long with strong for large model or large dataset. As the Hessian is a square matrix of size $d \times d$, the computation of Hessian in each client $k$, involves $\mathcal{O}(d^2)$ computation cost, $d$ is the number of model parameters, and computing the inverse of this Hessian associated with  $\mathcal{O}(d^3)$ computation cost and the memory requirement for storing Hessian is $\mathcal{O}(d^2)$. So, sustaining $\mathcal{O}(d^2) + \mathcal{O}(d^3)$ computation costs and $\mathcal{O}(d^2)$ memory costs is difficult for the local clients while dealing with large size model or large volume datasets \cite{singh2021nys,agarwal2017second}. Additionally, transmitting the Hessian in decentralized FL setups introduces significant communication overhead, making direct implementations impractical for resource-constrained edge devices. Apart from these, its not feasible to directly aggregate local models trained with Hessian matrix while updating the global model, as the local model involves Hessian inverse $\textbf{H}^{-1}_k$ and aggregation of these Hessian inverses does not satisfy the requirement of global Hessian inverse \cite{DDerezinskiM19} as shown below : 

\begin{equation}
    \textbf{H}^{-1} \neq \frac{1}{\mathcal{K}} \sum_{k = 1}^\mathcal{K} \textbf{H}^{-1}_k
    \label{Inverse:eq}
\end{equation}

These challenges necessitate the use of efficient approximations, such as diagonal, low-rank, and randomized methods, to make second-order optimization feasible in FL environments while retaining the benefits of curvature information. Approximation techniques, like the Nyström method or diagonal scaling, reduce the computational and communication demands associated with second-order methods while preserving their key advantages. These innovations make second-order optimization an increasingly practical choice for FL, addressing the limitations of first-order methods and enabling faster, more reliable convergence in distributed machine learning systems.

%%%%%%%%%%%%%%%%%%%%%%%%%%%%%%%%%%%%%%%%%%%%%%%%%%%%%%%%
%%%%%%%%%%%%%%%%%%%%%%%%%%%%%%%%%%%%%%%%%%%%%%%%%%%%%%
\section{Exploring Advanced Second-Order FL methods}
In this section, we categorize existing second-order FL methods based on how the Hessian curvature is utilized during global model updates in FL and discuss each method within the respective categories. Since the direct use of the Hessian matrix is computationally expensive, memory-intensive, and can lead to transition issues, research has addressed these challenges by approximating the Hessian curvature. Based on the method of Hessian curvature approximation, we categorize all existing methods into six groups: A. Hessian-free approach, B. Full Local Hessian Computation, C. Quasi-Newton based, D. Hessian diagonal approximation, E. Nyström method-based Hessian approximation and F. One-rank approximation of Hessian. With this categorization, we discuss each method under each category along with their empirical analysis, computation and memory costs, pros and cons. For local computation cost of each client, we consider per sample cost. 

\subsection{\textbf{Hessian-free approach}}
Hessian-free methods avoid the direct computation of the Hessian by using the Hessian-vector product \cite{martens2010deep,cho2015hessian}. Existing FL methods that use the Hessian-free approach include DANE \cite{shamirDANE_2014}, DiSCO \cite{zhang2015disco}, AIDE \cite{reddi2016aide}, FedDANE \cite{li2019feddane}, GIANT \cite{WangNeurIPSgiant_18}, DONE \cite{dinhdone_22}, LocalNewton \cite{guptalocalnewton_21} etc..

\subsubsection{\textbf{DANE}}
DANE is a Hessian-free method for second-order FL designed for strongly convex objectives, which directly computes the global model by averaging the locally trained models. To achieve this, DANE first computes the global gradient $\textbf{g}$ by averaging all the local gradients $\{\textbf{g}_k\}$. It then modifies each local objective by adding two regularization terms. The first term is based on the difference between the local gradient and the scaled global gradient (gradient correction term), while the second term is based on the difference between the local and global models (proximal term), as shown in Eq. \ref{Eq:5}. DANE then retrains the shared global model using this modified loss function. Local training with this modified objective incorporates curvature while updating both the local model and the aggregated global model. For quadratic loss functions, DANE achieves linear convergence.
\begin{equation}
    \min_{\textbf{x}} F'(\textbf{x}) =  f_k(\textbf{x}) - {(\textbf{g}_k - \eta\textbf{g})}^T \textbf{x}_k+\frac{\lambda}{2} {||\textbf{x}_k - \textbf{x}||_2^2}
    \label{Eq:5}
\end{equation}

\paragraph{Empirical Analysis} To validate the performance of DANE, the COV1 and ASTRO-PH \cite{shalev2013stochastic}, and MNIST \cite{lecun1998gradientmnist} datasets are used. For each dataset, the FL framework is created by randomly dividing the dataset into clients, and its performance is compared with one-shot parameter averaging \cite{zinkevich2010parallelized} and ADMM \cite{boyd2011distributed}. Experimental results on the regression problem show a faster reduction in training loss compared to ADMM and one-shot parameter averaging, in terms of fewer communication rounds.
\paragraph{Complexities}  
As DANE involves the computation of the global gradient and global Newton update separately, it requires four transmissions of $\mathcal{O}(d)$ from the server to the client and from the client to the server. The computation of the local gradient, along with updating the local model using the modified objective, incurs an overall $\mathcal{O}(d)$ computation cost and $\mathcal{O}(4d)$ memory cost. The overall server computation cost of DANE is $\mathcal{O}(\mathcal{K}d)$ for averaging local gradients and local models, and the overall memory cost is $\mathcal{O}(\mathcal{K}d + d)$ for storing local models and global gradient.

\paragraph{Pros}  
DANE is advantageous in terms of its ability to handle  distributed big-data regimes, improved convergence as compared to first-order methods in iid data and strongly convex objective, and its comparable local and server computation and memory costs relative to FedAvg.

\paragraph{Cons} It requires increased time to complete one communication round due to the four transmissions. Very limited samples in local clients can lead to non-convergence of the global model. Its not able to handle non-iid data across clients and partial participation of clients in each communication round. Its not applicable to non-convex optimization problems. It assumes equally distributed clients data while defining the objective function. Additionally, due to the computation of local newton update, the local memory costs is slightly higher than that of FedAvg.

\subsubsection{\textbf{AIDE}}
AIDE is an accelerated variant of DANE applicable to first-order oracles. It first modifies DANE by solving each local objective (as shown in Eq. \ref{Eq:5}) inexactly using algorithms like Quartz \cite{qu2015quartz}. Then, it incorporates Catalyst \cite{lin2015universal} into inexact DANE to accelerate convergence and make the process more communication-efficient.

\paragraph{Empirical Analysis} Experiments with federated binary classification using hinge loss on randomly and homogeneously partitioned RCv1, Covtype, Real-Sim, and URL datasets \cite{chang2011libsvm} demonstrate that AIDE outperforms COCOA+ \cite{takac2016distributed} and DANE in terms of faster reduction of function suboptimality.

\paragraph{Complexities}  
Similar to DANE.

\paragraph{Pros} AIDE is robust to bad data partitioning. It improves convergence as compared to DANE $\&$ COCOA+ and its local and server computation and memory costs are comparable  with FedAvg.

\paragraph{Cons} Same as DANE, it requires increased time to complete one communication round due to the four transmissions and applicable to strongly convex objective. Its not able to handle partial participation of clients in each communication round. Its not applicable to non-convex optimization problems. Due to the computation of local newton update, the local memory costs is slightly higher than that of FedAvg.

\subsubsection{\textbf{FedDANE}}
FedDANE is proposed to show the issues of non-IID data and partial participation of clients in DANE. To achieve this, FedDANE computes the global gradient using the gradients from a randomly selected subset of clients and computes the aggregated global model by solving the local subproblems used in DANE on another set of clients. Even FedDANE has convergence guarantees for both the convex and non-convex functions, empirical results shows an underperforming outcomes as compared to FedAvg. 

\paragraph{Empirical Analysis } Experiments are conducted on the FEMNIST image classification dataset with a convex model, the Shakespeare dataset for next-character prediction with a non-convex deep neural network (DNN) model, and the Sent140 dataset for sentiment analysis with a DNN model \cite{caldas2018leaf}. Each of these datasets is naturally partitioned among clients. Additionally, synthetic data with varying heterogeneity has been used for experiments. Findings from the experiments show that FedDANE performs worse in terms of reducing training loss when the data are heterogeneous across clients and when there is partial participation of clients. 

\paragraph{Complexities}  
Similar to DANE.

\paragraph{Pros} FedDANE is advantageous in terms of its ability to handle homogeneously distributed big-data regimes, improved convergence as compared to first-order methods in iid data and strongly convex objective, and its comparable local and server computation and memory costs relative to FedAvg.

\paragraph{Cons} Same as DANE, it requires increased time to complete one communication round due to the four transmissions. It under-performs first-order FL methods like FedAvg and FedProx \cite{li2020federatedprox} in non-iid data and partial client participation FL settings. Its not applicable to non-convex optimization problems. Additionally, due to the computation of local newton update, the local memory costs is slightly higher than that of FedAvg. 

\subsubsection{\textbf{DiSCO}} 
DiSCO is a FL algorithm designed for self-concordant and strongly convex local objective functions, which utilizes an inexact damped Newton method to compute the global model's update step, or the global Newton update. To this end, it employs the preconditioned conjugate gradient (PCG) method \cite{golub2013matrixPCG} with the global gradient while solving the regularized local subproblem, as shown in Eq. \ref{Eq:6A}, where $\lambda$ is the regularization parameter, to calculate the local Newton update. The local Newton updates are then aggregated to compute the global Newton update, which is used to update the global model. By utilizing PCG with the regularized local objective, DiSCO helps form a positive definite Hessian, making it invertible, and leads to super-linear convergence of the global model.

\begin{equation}
    \min_{\textbf{x}} f'_i(\textbf{x}) = f_i(\textbf{x}) + \frac{\lambda}{2} \|\textbf{x}\|_2^2
    \label{Eq:6A}
\end{equation}

\paragraph{Empirical Analysis } The experiments are conducted on federated binary classification using logistic regression on three LIBSVM datasets \cite{chang2011libsvm} (i.e., Covtype, RCV1, and News20) to validate the effectiveness of DiSCO and compare its performance with ADMM \cite{boyd2011distributed}, Accelerated Full Gradient (AFG) \cite{nesterov2013introductory}, L-BFGS \cite{liu1989limited}, and DANE \cite{shamirDANE_2014}. For each dataset, the client distribution is created using random partitions. Experimental results show the superior performance of DiSCO compared to the other methods in terms of faster reduction in training loss.

\paragraph{Complexities}  DiSCO requires sending the initial global model of size $\mathcal{O}(d)$ to each client. Each client then broadcasts its local gradient of size $\mathcal{O}(d)$ to the server. The server subsequently broadcasts the aggregated global gradient along with a scaling vector $\textbf{u}$, resulting in an overall transmission load of $\mathcal{O}(2d)$. Each client computes its local Newton update of size $\mathcal{O}(d)$ and shares it with the server. Therefore, completing all these processes requires four transmissions per communication round. DiSCO incurs an overall $\mathcal{O}(d)$ computation cost and $\mathcal{O}(3d)$ memory cost at each client (local model + local gradient + Newton update). The overall server computation cost for DiSCO, which involves computing the global gradient and the global Newton update using PCG, is $\mathcal{O}(\mathcal{K}d)$, and the overall memory cost is $\mathcal{O}(d + \mathcal{K}d)$.

\paragraph{Pros} DiSCO is advantageous in terms of its super-linear convergence, and its comparable local and server computation and memory costs relative to FedAvg.

\paragraph{Cons} The disadvantages of DiSCO are as follows: it is designed for self-concordant and strongly convex local objective. It also requires increased time to complete one communication round due to the four transmissions. Additionally, it assumes that the number of samples is the same across all clients. Its not applicable to non-convex optimization problems. It requires twice differentiable objective function. Additionally, due to the computation of local newton update, the local memory costs is slightly higher than that of FedAvg.

\subsubsection{\textbf{GIANT}}
GIANT approximates the local Newton update using $m$ conjugate gradient (CG) iterations \cite{nazareth2009conjugate} to solve a strongly convex and twice differentiable objective function as given in Eq. \ref{Eq:6}, where $\lambda$ is regularization parameter of the global objective function.  

\begin{equation}
    \min_{\textbf{x}} F(\textbf{x}) = \frac{1}{\mathcal{K}} \sum_{k \in \mathcal{K}} f_k(\textbf{x}) + \frac{\lambda}{2} {||\textbf{x}||_2^2}
    \label{Eq:6}
\end{equation}
While applying the CG method for calculating the local Newton update, GIANT uses the global gradient computed by averaging all the local gradients. Once all the local Newton updates, $\{\Delta_k\}$, are computed, GIANT aggregates them to compute the global Newton update and uses it to update the global model. In computing the global Newton update, GIANT approximates the global Hessian (i.e., the arithmetic mean of local Hessians) by using the harmonic mean of the local Hessians, under the assumption that the data are homogeneous across clients. The use of the CG-based Newton update computation in GIANT leads to linear-quadratic convergence of the global model, which is advantageous in homogeneously distributed big-data regimes.

\paragraph{Empirical Analysis } The performance of GIANT is validated through experiments on randomly partitioned MNIST8M, Epsilon and Covtype datasets available in LIBSVM website, where $20\%$ data is used for checking the performance of the global model. The federated classification tasks for each of these datasets is performed using logistic regression. The outcomes from the experiments shows that GIANT outperforms accelerated gradient descent (AGD) \cite{nesterov2013introductory}, L-BFGS \cite{liu1989limited}, and DANE \cite{shamirDANE_2014} in terms of reducing classification error. \\

\paragraph{Complexities}  
As GIANT involves the computation of the global gradient and global Newton update separately, it requires four transmissions of $\mathcal{O}(d)$ from the server to the client and from the client to the server. To compute the local gradient and local Newton update using $m$ iterations of CG, GIANT incurs an overall $\mathcal{O}(d)$ computation cost and $\mathcal{O}(4d)$ memory cost at each client (local model + local gradient + global gradient + auxiliary variables). The overall server computation cost of GIANT is $\mathcal{O}(\mathcal{K}d)$ for averaging local gradients and local Newton updates, and the overall memory cost is $\mathcal{O}(d + \mathcal{K}d)$ (previous global model + local models).

\paragraph{Pros} GIANT is advantageous in terms of its ability to handle big-data regimes, its linear-quadratic convergence, and its comparable local and server computation and memory costs relative to FedAvg.

\paragraph{Cons} GIANT is designed for a twice-differentiable and strongly convex global objective with homogeneous data across clients. It also requires increased time to complete one communication round due to the four transmissions. Its not applicable to non-convex optimization problems. It is assumed that each of client is having equal number of samples while designing the objective of GIANT. Additionally, due to the computation of local newton update, the local memory costs is slightly higher than that of FedAvg.

\subsubsection{\textbf{DONE}}
DONE is designed for improving convergence of the global model in FL with strongly convex objective function (Eq. \ref{Eq:6}). To achieve the same, DONE computes local Newton update using $m$ Richardson iterations \cite{rheinboldt2009classical} along with the global gradient and aggregate them to compute global Newton update. Similar to GIANT, DONE also requires four times transmission in each communication round between server and each client. Due to use of Richardson iterations, DONE achieve linear-quadratic convergence and it can be applicable to the non-iid data across clients.  
\paragraph{Empirical Analysis } The performance of DONE is validated through experiments on MNIST \cite{lecun1998gradientmnist}, FEMNIST \cite{cohen2017emnist}, and Human Activity Recognition (HAR) \cite{anguita2013publicHAR} datasets with non-iid FL settings. For MNIST and FEMNIST, data are partitioned by dividing samples from certain number of classes to each and every clients. For HAR dataset, data collected from each individual is considered as a client. For federated classification task of each of these dataset, logiostic regression model is used. Findings from the experiments demostrate that DONE performs well as compared to FedAvg and Newton-based methods such that GIANT, FEDL, and DANE in terms of lower training loss and better test accuracy. 
\paragraph{Complexities}  
The complexities of DONE are similar to GIANT.
\paragraph{Pros} DONE is advantageous in terms of its ability to work well in non-iid data, its linear-quadratic convergence, and its comparable local and server computation and memory costs relative to FedAvg.

\paragraph{Cons} DONE is designed for a twice-differentiable and strongly convex global objective with homogeneous data across clients. It also requires increased time to complete one communication round due to the four transmissions. Its not applicable to non-convex optimization problems. Additionally, due to the computation of local newton update, the local memory costs is slightly higher than that of FedAvg.

\subsubsection{\textbf{LocalNewton }}
LocalNewton can be viewed as an improvement of GIANT in terms of reduced communication between the server and clients. LocalNewton performs multiple local iterations based on the conjugate gradient (CG) method with local gradients, while updating the local model and performing aggregation of local models following GIANT once the local models are updated. This process reduces the frequent transmission overhead encountered by GIANT and improves model convergence. Basically it reduce number of communication rounds by increasing number of local iterations. Its simplicity, reduced communication costs, and scalability make it well-suited for applications in distributed machine learning, federated learning, and serverless systems. However, due to multiple local iterations, LocalNewton increases local computation costs. The algorithm introduces an adaptive mechanism to dynamically adjust the number of local iterations, further optimizing communication efficiency and computational performance. Additionally, LocalNewton serves as an initialization scheme that brings model estimates closer to the optimal solution, after which a fully synchronized second-order method, such as GIANT, can be employed to refine convergence. 

\paragraph{Empirical Analysis } The performance of LocalNewton is validated through experiments on randomly and equal partitioned W8a, Epsilon, a9a, ijcnn1 and Covtype datasets available in LIBSVM website. The federated classification tasks for each of these datasets is performed using logistic regression. The outcomes from the experiments shows that LocalNewton outperforms GIANT, L-BFGS, and Local SGD in terms of reducing training loss and achieving better test accuracy. 

\paragraph{Complexities} The per-iteration local computation time of LocalNewton is \( r \) times higher than that of GIANT due to performing \( r \) local iterations instead of one iteration, even though the overall local computation cost is the same as GIANT, i.e., \( \mathcal{O}(d) \). Its local memory cost is \( \mathcal{O}(d) \) less than GIANT as there is no requirement for storing the global gradient. The server computation cost for aggregating local models is the same as GIANT, but the server memory cost is \( \mathcal{O}(d) \) lower than GIANT due to not requiring the storage of the previous global model in the server, as the server simply takes the averages of local models to update the global model. The transmission cost of LocalNewton is the same as FedAvg and only requires two transmissions per communication round.
\paragraph{Pros} LocalNewton reduces communication rounds by performing multiple local updates before synchronizing with the master, making it well-suited for high-latency environments such as federated learning and serverless systems. Its adaptive version dynamically adjusts the number of local iterations, enhancing model updates and reducing redundant computations as training progresses. Empirical evaluations demonstrate significant reductions in both training times and communication rounds compared to state-of-the-art methods in real-world distributed environments.
\paragraph{Cons} LocalNewton may exhibit an error floor in non-convex settings or when local data is limited, preventing it from converging to the exact solution. Its performance can also degrade with non-uniform data partitioning, which increases gradient variance and slows convergence. Additionally, while reducing communication costs, LocalNewton imposes a higher computational burden on clients due to local second-order updates, potentially limiting its applicability on resource-constrained devices. Finally, its theoretical guarantees depend on strong assumptions such as strong convexity and smoothness, which may not hold in all practical scenarios. Additionally, due to the computation of local newton update, the local memory costs is slightly higher than that of FedAvg.

\subsection{\textbf{Full Local Hessian Computation}}
This methods involves with the calculation of explicitly calculation of Hessian in local clients and sending the the Hessian information in a compressed form to avoid excessive transmission cost of Hessian. Exiting methods include FedNL \cite{safaryanfednl_22}, Basis Matters \cite{qianbasis_matter_21}, FedNew \cite{elgabli_fednew_22}, SHED \cite{dal2024shed}, FedNS \cite{li2024fedns} etc.. 

\subsubsection{\textbf{FedNL}}
FedNL is proposed as the modification of the novel distributed learning learning algorithm namely Newton Learn \cite{NL1}, which overcome the issues of data privacy and applicability of Newton learn by introducing an innovative Hessian learning mechanism that learns the next step's the Hessian approximation by using its past immediately past Hessian for both the clients (Eq. \ref{NL:Eq:1}) and the server (Eq. \ref{NL:Eq:2}) along with the utilization of more aggressive contractive compression of the information before interchanging between the server and the clients. The approximation of the local Hessian using Eq. \ref{NL:Eq:1} is robust to the application of contractive compression like Top-K or Rank-R for reducing the Hessian transmission cost from the client to the server from $\mathcal{O}(d^2)$ to $\mathcal{O}(d)$ and enjoys linear (Lyapunov function) or super linear (distance to the optima) convergence independent to the condition number.
 
\begin{equation}
    \textbf{H}_k^{t+1} = \textbf{H}_k^{t} + \alpha \textbf{S}_k^t, \quad
    \mathbf{S}^t_k = C_k^t \left( \nabla^2 f_k(\textbf{x}^t) - \mathbf{H}_k^t \right)  
    \label{NL:Eq:1}
\end{equation}
where \( C_k^t \) is a compression operator and \( \alpha > 0 \) is the Hessian learning rate.
\begin{equation}
    \textbf{H}^{t+1} = \textbf{H}^{t} + \alpha \textbf{S}^t, \quad \mathbf{H}^0 = \frac{1}{\mathcal{K}} \sum_{i=1}^\mathcal{K} \mathbf{H}_k^0,  \quad \textbf{S}^t =\frac{1}{\mathcal{K}} \sum_{i=1}^\mathcal{K} \mathbf{S}_k^t
    \label{NL:Eq:2}
\end{equation}

\paragraph{Empirical Analysis } The performance of FedNL is validated through experiments on randomly and equally partitioned a1a, w7a, w9a, w8a, and phishing datasets available in LIBSVM. The federated binary classification tasks for each of these datasets is performed using regularized logistic regression. The outcomes from the experiments shows that FedNL outperforms vanilla gradient descent, DIANA \cite{DIANA_2024}, ADIANA \cite{ADIANA_li_20}, shifted local gradient descent \cite{gorbunov2021local}, and DINGO \cite{crane2019dingo} in terms of lower classification error while minimizing transmission overhead.

\paragraph{Complexities} The approximation and compression of local Hessian is associated with handling a square matrix of size $d$. So overall local computation cost is $\mathcal{O} (d^2)$ and local memory requirement for storing Hessian, gradient and local model is $\mathcal{O} (d^2 + 2d)$. For aggregating compressed Hessian, aggregating local gradients, decoding  the aggregated compressed Hessian and to inverse this, it takes overall $\mathcal{O} (d^3)$ computation costs and $\mathcal{O} (d^2 + 2d)$ memory costs (for storing full Hessian, global gradient and previous global model). The transmission costs for compressed Hessian along with the gradient from the client to the server is $\mathcal{O}(r + d)$ and from the server to the client, its incurs $\mathcal{O}(r)$ costs, where $r<d$ is the size of bidirectional compression of gradient and the global model used in FedNL. 

\paragraph{Pros} FedNL is generally applicable to a wide range of finite-sum problems, supporting heterogeneous data distributions and partial client participation without imposing local data to the server. Its support for unbiased, contractive compression strategies like Top-K and Rank-R significantly reduces communication costs. Moreover, FedNL achieves fast local convergence with a linear rate that remains independent of the condition number, data volume, or compression variance.

\paragraph{Cons} The central server as well as local clients bear a significant computational burden when processing compressed Hessians, while the method's performance is highly sensitive to hyperparameters such as the Hessian learning rate ($\alpha$) and the specifics of the compression operators ($C_{ki}$). Moreover, although it supports both unbiased and contractive compressors, its performance can degrade under high compression variance.

\subsubsection{\textbf{Basis Matters}}
Basis Matters is simple the generalization of FedNL. Like FedNL, Basis Matters computes local Hessian using previous Hessian in the local client as well as the server. But the only difference is that it changes the representation of the local Hessian by learning the basis of this Hessian and then apply compression mechanism on it before sending it to server, which helps to further reduce the transmission cost from $\mathcal{O}(d)$ to $\mathcal{O}({r'}^2)$, where, $r' << d$.

\paragraph{Empirical Analysis } a1a, a9a, phishing, w2a, w8a, covtype, madelon of LIBSVM are used and for experiments with similar partition of FedNL. The outcomes shows that Basis Matters outperforms vanilla gradient descent, DIANA \cite{DIANA_2024}, ADIANA \cite{ADIANA_li_20}, and FedNL in terms of lower classification error while minimizing transmission overhead.

\paragraph{Complexities} The server and local computation and memory costs are similar to those of FedNL. Due to the use of the basis of the Hessian, the transmission costs from client to server further reduce to $\mathcal{O}({r'}^2)$, where $r'<<d$ as compared to FedNL, while server to client transmission is the same as FedNL if they use the same compression techniques. 

\paragraph{Pros} Same as FedNL with one additional advantage of further reducing the transmission costs from the client to the server. 

\paragraph{Cons} Same as FedNL.

\subsubsection{\textbf{FedNew}}
FedNew introduces a novel second-order optimization framework for federated learning, addressing high transmission costs and privacy concerns caused by sharing Hessian and gradient with the server. It computes Hessian ($\textbf{H}_k$) and gradient ($\textbf{g}_k$) in each local client and finds the inverse Hessian and gradient product ($ \Delta_k = \textbf{H}^{-1}_k \textbf{g}_k$) using a single step of the alternating direction method of multipliers (ADMM) by optimizing Eq. 10 of \cite{elgabli_fednew_22}. The host server collects and aggregates all the local updates $\{\Delta_k \}$ and computes the global update $\Delta$, which is used for updating the global model. Additionally, it incorporates quantization of local updates while sending them to the server, which further reduces local transmission costs. As FedNew is computing the Hessian matrix locally, it increases local computation costs as well as the aggregation of local Newton updates incurs the global Hessian inverse issue mentioned in Eq. \ref{Inverse:eq}. FedNew is primarily applicable to strongly convex objectives due to its reliance on ADMM for local updates. For non-convex objectives, FedNew may still be applied, but the theoretical guarantees are less clear, and performance might not be as robust without further adaptations or guarantees in the non-convex setting.

\paragraph{Empirical Analysis } The performance of FedNew is validated through experiments on evenly and equally partitioned a1a, w7a, w8a, and phishing datasets available in LIBSVM. The federated binary classification tasks for each of these datasets is performed using regularized logistic regression (as given in Eq. \ref{Eq:6}). The outcomes from the experiments shows that FedNew outperforms FedAvg and FedNL \cite{safaryanfednl_22} in terms of reducing number of communication rounds. 

\paragraph{Complexities} The overall  per-iteration local computation cost for calculating inverse of local Hessian, local gradient and local Newton update is $\mathcal{O} (d^3)$ and memory cost is $\mathcal{O} (d^2 + 2d)$ for storing the Hessian, gradient and local model. The overall server computation cost is $\mathcal{O}(\mathcal{K}d)$ for averaging local updates, and the overall memory cost is $\mathcal{O}(d + \mathcal{K}d)$ (previous global model + local models). Due to sharing quantized form of local Newton update, it has $\mathcal{O} (r)$ ($r < d$) client to server transmission cost, which is less than FedAvg and the server to client transmission cost is same as FedAvg. 

\paragraph{Pros } Sharing the quantized Newton update, i.e. quantized inverse Hessian and gradient product, from the client to server leads to preserving privacy of the local data and lower transmission costs.

\paragraph{Cons } The calculation of the local Hessian increases local computation and memory costs, and the aggregation of local Newton updates while finding the global Newton update incurs the global Hessian inverse issue mentioned in Eq. \ref{Inverse:eq}. Although empirical convergence has been observed, theoretical guarantees are currently limited to asymptotic settings, leaving non-asymptotic analysis as an open challenge.

\subsubsection{\textbf{SHED}}
SHED is a second-order FL method designed to handle non-iid data, heterogeneity in transmission resources, and to reduce communication rounds in FL. To this end, SHED computes the exact Hessian (\( \mathbf{H}_k \)) at the local client and performs eigenvalue decomposition using Eq. \ref{Eq:eig}, where \( \mathbf{V}_k \) represents the eigenvectors and \( \Lambda_k \) denotes the eigenvalues of the Hessian.

\begin{equation}
    \mathbf{H}_k = \mathbf{V}_k \Lambda_k \mathbf{V}_k^T
    \label{Eq:eig}
\end{equation}

Then, SHED shares some of the eigenvalues and their corresponding eigenvectors, along with the local gradient, with the server. While sending these, it considers the eigenvalues in decreasing order. Once the server receives this local information, it approximates the local Hessian using the eigenvectors and aggregates the local Hessians along with the local gradients. These are then used to compute the Newton update. When approximating the local Hessian, SHED employs an incremental strategy to select a set of eigenvalues in each communication round, such that at a certain round, all the eigenvalues are seen, and the approximation of the local Hessian becomes a full-rank Hessian. For general convex problem, where the Hessian changes based on model parameters, SHED recompute the approximation of the local Hessian in a certain interval of communication rounds instead of computing once. 

\paragraph{Empirical Analysis } The performance of SHED is examined through experiments on Million Song, FMNIST, EMNIST and w8a datasets. The federated classification tasks for each of these datasets is performed using logistic regression. The outcomes from the experiments in both the iid and non-iid show that SHED outperforms BFGS, GIANT and FedNL in terms of reducing classification errors in comparatively lower number of communication rounds.

\paragraph{Complexities } Due to the computation of the local gradient, local Hessian, and its eigenvalue decomposition, SHED has an overall local computation cost of \( \mathcal{O}(d^2) \) and memory cost of \( \mathcal{O}(2d + d^2) \) (due to storing of local model, local gradient and local Hessian). As the server approximates the local Hessian, it results in \( \mathcal{O}(d^2) \) computation. Aggregating the local Hessians and local gradients results in \( \mathcal{O}(\mathcal{K} d^2) \) server computation, with \( \mathcal{O}(d + \mathcal{K} d^2 + \mathcal{K} d) \) memory requirements. Since SHED sends a set of \( r' \) eigenvectors along with the gradient, the transmission cost from the client to the server is \( \mathcal{O}(r'd + d) \), and the server-to-client transmission cost is \( \mathcal{O}(d) \), as it only sends the model parameters from the server.

\paragraph{Pros} SHED can handle heterogeneous data distribution among clients. It can control the transmission cost from the client to the server based on the communication resources available at the client. It achieves super-linear convergence of the global model for convex objective functions.

\paragraph{Cons} The main disadvantage of SHED is the high computation and memory costs encountered by both the clients and the server due to the computation and storage of the Hessian. SHED is designed under the assumption of an equal number of samples at each client, which is not always feasible in real-life applications.

\subsubsection{\textbf{FedNS}}
FedNS aims to accelerate FL training to super-linear convergence while reducing the transmission cost associated with the Hessian. To this end, it computes the local Hessian and performs Newton sketching on the square root of the Hessian ($\textbf{H}^{\frac{1}{2}} \in \mathbb{R}^{n_k \times d}$) to reduce the Hessian size before transmitting it to the server. While performing Newton sketching, it uses a sketch matrix $\textbf{S}_k \in \mathbb{R}^{r \times n_k}$ with zero mean and $\mathbb{E}[\textbf{S}_k \textbf{S}_k^T] = \textbf{I}_{n_k}$, where $n_k$ is the number of samples in client $k$. The computation of the sketched Hessian ($\textbf{Y}_k \in \mathbb{R}^{r \times d}$) is done through Eq. \ref{HesSket:eq}.

\begin{equation}
    \textbf{Y}_k = \textbf{S}_k \textbf{H}^{\frac{1}{2}}
    \label{HesSket:eq}
\end{equation}

The server then collects the sketched local Hessian along with the local gradient from each client and approximates the full Hessian using Eq. \ref{aprox_hessin:eq}, where $\lambda$ is a regularization parameter. Once the approximation of all the local Hessians is done, the server aggregates them to compute an approximation of the global Hessian and uses this Hessian along with the global gradient to update the global model using second-order optimization. 

\begin{equation}
    \widetilde{\textbf{H}}_k = {\textbf{Y}_k}^T \textbf{Y}_k + \lambda \textbf{I}
    \label{aprox_hessin:eq}
\end{equation}

\paragraph{Empirical Analysis } The performance of FedNS is examined through experiments on evenly and equally partitioned phishig, cod-rna, covtype and SUSY datasets available in LIBSVM. The federated classification tasks for each of these datasets is performed using logistic regression. The outcomes from the experiments show that FedNS outperforms FedAvg, FedNewton, FedNew and FedNL in terms of reducing classification errors in comparatively lower number of communication rounds. 

\paragraph{Complexities} The overall  per-iteration local computation cost of FedNS is $\mathcal{O} (d^2)$ as it is associated with the calculation and  sketching of a regularized Hessian, and memory cost is $\mathcal{O} (d^2 + 2d)$ for storing the Hessian, gradient and local model. The server computation cost is $\mathcal{O}(\mathcal{K}d^2)$ for averaging local Hessians $\&$ local gradients and for inverting the global Hessian, it requires $\mathcal{O}(d^3)$. So, the overall server computation $\mathcal{O}(d^3)$, when $\mathcal{K} < d$, otherwise $\mathcal{O}(\mathcal{K}d^2)$. The memory requirements of FedNS in server is $\mathcal{O}(\mathcal{K}d^2 + \mathcal{K}d + d)$ for storing local Hessians, local gradients and previous global model. Due to sharing sketched Hessian along with gradient, client to server transmission cost is $\mathcal{O}(rd + d)$ and server to client cost is $\mathcal{O}(d)$.

\paragraph{Pros } Sharing the sketched Hessian, from the client to server leads to super-linear convergence,  preserving privacy of the local data and lower transmission costs as compared to that of sending exact Hessian. 

\paragraph{Cons } Calculation and storing of Hessian, increase computation and memory load in both the client and server.

\subsection{\textbf{Quasi-Newton Based}}
Quasi-Newton method \cite{nocedal2006stephenn_numerical_optimization} is a variant of second-order optimization where the Hessian is approximated using only first-order information via secant equation as shown in Eq. \ref{Eq:secant}, where $\textbf{B}^t$ is the approximation of Hessian, $\textbf{s}^t = \textbf{x}^{t+1} - \textbf{x}^{t}$ is difference between current and previous model parameters and $\textbf{y}^t = \textbf{g}^{t+1} - \textbf{g}^{t}$ is difference between current and previous gradients. Exiting FL methods based on Quasi-Newton approach include FedSSO \cite{Mafedsso_22} and DQN-Fed \cite{hamidi2025distributed}.

\begin{equation}
    \textbf{B}^{t+1} \textbf{s}^t = \textbf{y}^t
    \label{Eq:secant}
\end{equation}

\subsubsection{\textbf{FedSSO}}
FedSSO represents a significant advancement in second-order optimization for FL, designed to overcome the limitations of client-side second-order computations by shifting the computational burden entirely to the server. Unlike traditional methods that require intensive client-side computations and frequent communication of gradients and Hessians, FedSSO employs a server-side Quasi-Newton approximation of the global Hessian, eliminating communication of second-order updates and reducing overall communication costs. The global Hessian matrix is approximated on the server using the BFGS \cite{nocedal2006stephenn_numerical_optimization} algorithm as follows :

\[
\textbf{B}^{t+1} = \textbf{B}^{t} + \frac{\textbf{y}^{t} {\textbf{y}^{t}}^T}{{\textbf{y}^{t}}^T \textbf{s}^{t}} 
- \frac{\textbf{B}^{t} \textbf{s}^{t} {\textbf{s}^{t}}^T \textbf{B}^{t}}{{\textbf{s}^{t}}^T \textbf{B}^{t} \textbf{s}^{t}}
\]

\paragraph{Empirical Analysis }
The emperical evaluation of FedSSO is conducted on heterogeneously partitioned MNIST, EMNIST, CIFAR10 \cite{krizhevsky2009learning_cifar10} and LIBSVM datasets. For each dataset, non-iid partition is made by considering number of samples, distribution of data categories, and the data categories on each client. The experiments are conducted both convex (with logistic regression model) and non-convex (with CNN) settings. Results show that FedSSO performs well that first order methods like FedSGD, FedAvg, , FedAC \cite{yuan2020federated_AC}, FedOpt \cite{reddi2020adaptive_opt}, SCAFFOLD \cite{KarimireddySCAFFOLD_20} and second-order methods such that FedDANE \cite{lifeddane_19} and FedNL in terms of train loss and test accuracy.

\paragraph{Complexities} The local computation costs is $\mathcal{O}(d)$ for computing gradient and memory costs is $\mathcal{O}(2d)$ (for storing local model and gradient), and the server computation cost for averaging local gradients is $\mathcal{O}(\mathcal{K}d)$. The computation of the Newton update with the approximated Hessian has a cost of $\mathcal{O}(d^2)$, so the overall server computation cost is $\mathcal{O}(d^2)$. The server memory cost is $\mathcal{O}(4d + d^2)$ due to storing of global gradient, $\textbf{y}^t$, $\textbf{s}^t$, previous global model and the approximated global Hessian. Both server to client and client to server transmission costs is $\mathcal{O}(d)$ as FedSSO requires to share local gradient from client and model parameters from the server.

\paragraph{Pros} FedSSO shifts the computation of second-order updates to the server, eliminating the need for exchanging Hessian terms between clients and the server and thereby significantly reducing communication costs. Moreover, theoretical analysis confirms that FedSSO achieves a sub-linear convergence rate of $\mathcal{O}\left(\frac{1}{k}\right)$ for convex settings, comparable to existing stochastic Quasi-Newton methods. In addition, FedSSO demonstrates superior adaptability by delivering improved convergence in both convex and non-convex scenarios under heterogeneous (non-IID) data distributions, outperforming traditional methods such as FedAvg and SCAFFOLD in experimental evaluations.

\paragraph{Cons} FedSSO reduces client-side computational costs by shifting the heavy burden of Hessian approximation and global model updates to the server; however, this increases the computational demand on the server, which may require substantial resources. Additionally, initial training phases can exhibit instability in both convex and non-convex models, particularly when starting from a model far from optimal, and the centralized server approach may encounter scalability challenges as the number of clients grows substantially, potentially leading to bottlenecks.

\subsubsection{\textbf{DQN-Fed}}
DQN-Fed addresses the issue of unfair aggregation of local models caused by heterogeneous data across clients in FL by ensuring that all local losses decrease and that the rate of change of local losses aligns with the quasi-Newton method. It also ensures that the global model converges at the Pareto-stationary point \cite{mukai2003algorithms} of the global objective. The rate of change of each local loss, denoted by $d^t_k$, is computed via the first-order approximation of the loss function at the updated point $x^{t+1}$, as given in Eq. \ref{Eq:rate_grads}, where the inverse of the Hessian is approximated using the BFGS algorithm to reduce the excessive computational load from $\mathcal{O}(d^2 + d^3)$ to $\mathcal{O}(d^2)$.

\begin{equation}
    d^t_k = {\textbf{g}_k^t}^T {\textbf{H}_k^t}^{-1} \textbf{g}_k^t
    \label{Eq:rate_grads}
\end{equation}

To achieve the aforementioned goal, DQN-Fed first orthogonalizes the local gradients using a modified Gram–Schmidt orthogonalization process by incorporating \{$d^t_k$\} as given in Eq. 8 of \cite{hamidi2025distributed}. These orthogonalized gradients are then aggregated with the goal of finding the minimum-norm vector in the convex hull of these orthogonal gradients to form the global Newton update. The aggregation rule is shown in Eq. 12 of \cite{hamidi2025distributed}.
 
\paragraph{Empirical Analysis } The experiments are conducted on heterogeneously partitioned CIFAR10, CIFAR100, Shakespeare, FMNIST, Tiny ImageNet \cite{le2015tiny}, and CINIC10 \cite{darlow2018cinic} datasets. The heterogeneity across clients is produced by using two types of division strategies: 1. Based on class labels (with equal size of local samples) \cite{wang2021federated}, 2. Dirichlet distribution-based with a concentration parameter of 0.5 (unequal sample size) \cite{wang2020federated}. In the first setting, a feedforward neural network with 2 hidden layers is used, and in the second setting, a ResNet-18 model \cite{he2016deep} is used. The results demonstrate a sharp improvement in achieving better test accuracy faster compared to first-order fair FL methods such as qFFL, TERM, FedMGDA, Ditto, FedLF, and FedHEAL (descriptions of these algorithms are provided in Section 5 of \cite{hamidi2025distributed}), along with second-order FL methods like FedNL and FedNew.
 
\paragraph{Complexities} As DQN-Fed involves the approximation of the local Hessian inverse via BFGS, it is associated with an overall local computation cost of $\mathcal{O}(d^2)$. The local memory cost is $\mathcal{O}(4d + d^2)$. The server computation is $\mathcal{O}(\mathcal{K}d)$ for the orthogonalization and aggregation of orthogonal gradients and the overall memory costs is $\mathcal{O}(d + 2\mathcal{K}d)$ for storing both local gradients and orthogonal gradients along with the past global model. Same as FedSSO, Both server to client and client to server transmission costs is $\mathcal{O}(d)$ DQN-Fed. 

\paragraph{Pros} DQN-Fed achieves linear-quadratic convergence rate while maintaining fair aggregation of the local models there by improve FL model training in heterogeneous FL settings. 

\paragraph{Cons} DQN-Fed increases client-side computational and memory requirements due to calculation of Hessian inverse locally. Additionally, initial training phases can exhibit instability in both convex and non-convex models, particularly when starting from a model far from optimal.

\subsection{\textbf{Hessian diagonal approximation}}
To avoid excessive computation and memory overhead with full Hessian matrix, another way of utilizing curvature information is directed through the approximation of diagonal Hessian. Exiting second-order FL method that utilizes Hessian diagonal while updating the global model includes FopLAHD \cite{senfoplahd_23}, linear-time diagonal approximation (LTDA) \cite{senpreml_23}, HWA \cite{ahmad2023robust}, and Fed-Sophia \cite{elbakary2024fed}. 

\subsubsection{\textbf{FopLAHD}} 
The computation of the Hessian diagonal involves the same computations as the full Hessian but is more memory efficient. To reduce this computation, FopLAHD approximates the diagonal using the Hessian-vector product as described in \cite{bekas2007estimator}, using $r \ll d$ iterations. FopLAHD reduces the variance induced in the approximated diagonal vector, detects the noise, and replaces it with the mean value $e_m$. When detecting the noise, it uses a condition based on lower and upper thresholds on the mean value, i.e., if any element $e < \tau e_m$ or $e > e_m + (1-\tau) e_m$, replace $e$ with $e_m$, where $\tau \in (0, 1]$. FopLAHD aggregates all the local Hessian diagonals along with local gradients and updates the global model by replacing the Hessian with the diagonal vector in the second-order optimization equation as shown in Eq. \ref{diag:eq}, where $\rho$ is the diagonal regularization term that helps tackle the inversion issue caused by any element found to be zero in the diagonal vector. Additionally, it uses the momentum of the global gradient and global Hessian to tackle partial client participation.
\begin{equation}
    \textbf{x}^t = \textbf{x}^{t-1} - {(\textbf{Z}^t + \rho)}^{-1} \textbf{g}^t
    \label{diag:eq}
\end{equation}

\paragraph{Empirical Analysis } Experiments on heterogeneously partitioned and partial client federated image classification on MNIST and FMNIST datasets with a logistic regression model demonstrate that FopLAHD outperforms FedProx, SCAFFOLD, DONE, and GIANT in terms of reducing training and test losses, as well as achieving better test accuracy in fewer communication rounds. The heterogeneity across clients is created using Dirichlet distribution with very low concentration parameter (0.2).

\paragraph{Complexities} For approximating local Hessian diagonal and gradient, FopLADH requires overall $\mathcal{O}(d)$ computation costs and overall memory costs  $\mathcal{O}(4d)$ in local clients for storing local model, gradient and auxiliary vectors $\textbf{p}_k$ and $\textbf{q}_k$. In server, aggregation of local diagonal and local gradients requires overall $\mathcal{O}(\mathcal{K}d)$ computation costs. For storing all the local diagonals, local gradients and previous global model along with the moments of global gradient and global Hessian diagonal, server needs overall $\mathcal{O}(2\mathcal{K}d + 3d)$ memory. Server to client transmission costs of FopLADH is $\mathcal{O}(d)$ and client to server is $\mathcal{O}(2d)$.

\paragraph{Pros} FopLADH Reduces local computation for Hessian computation from $\mathcal{O}(d^2)$ to $\mathcal{O}(d)$ ($r<<d$). It can perform well on heterogeneous and partial client FL. 

\paragraph{Cons}  It is limited by its reliance on a diagonal approximation, which ignores off-diagonal elements that may capture important curvature details in highly non-convex settings. Additionally, each client must compute and share both gradients and Hessian diagonals, slightly increasing computational and communication overhead compared to FedAvg. Need twice differentiable objective function to approximate the Hessian diagonal. 

\subsubsection{\textbf{Linear-Time Diagonal Approximation (LTDA)}}
In this paper, the local Hessian diagonal is approximated using the first row of the true Hessian and uses a regularization term with the approximated Hessian while updating the local model in each client. The computation of first row ($\textbf{h}_1$) is shown in Eq. \ref{first_hessin:eq}, where $\textbf{g}^t_k [0]$ is the first element of the gradient $\textbf{g}^t_k$ and the approximation of the Hessian diagonal is shown in Eq. \ref{approx_hessin:eq}, where $\textbf{h}_1 [0]$ is the first element of the first row $\textbf{h}_1$.

\begin{equation}
    \textbf{h}_1 = \frac{\partial \textbf{g}^t_k [0]}{\textbf{x}^t_k}
    \label{first_hessin:eq}
\end{equation}

\begin{equation}
    \textbf{Z}^t = \frac{\textbf{h}_1 \cdot \textbf{h}_1}{\textbf{h}_1 [0]}
    \label{approx_hessin:eq}
\end{equation}

Once the local model is updated using Hessian diagonal and local gradient, it is then shared with the server and the server aggregates all the local models to update the global model. 

\paragraph{Empirical Analysis } Experiments on heterogeneously partitioned and partial client federated image classification on MNIST and FMNIST datasets with a logistic regression model demonstrate that the proposed method outperforms FedAvg, FedProx, SCAFFOLD, and DONE in terms of reducing training and test losses, as well as achieving better test accuracy in fewer communication rounds. The data partitions across clients are made by considering class labels, similar to \cite{wang2021federated}.
  
\paragraph{Complexities} For approximating local Hessian diagonal and gradient, it requires overall $\mathcal{O}(d)$ computation costs and overall memory costs  $\mathcal{O}(3d)$ in local clients for stroring local model, gradient and Hessian's diagonal vector. In server, aggregation of local models requires overall $\mathcal{O}(\mathcal{K}d)$ computation costs. For storing all the local models along with previous global model, server needs overall $\mathcal{O}(\mathcal{K}d + d)$ memory. Server to client transmission costs is $\mathcal{O}(d)$ and client to server is $\mathcal{O}(d)$, which is same as FedAvg.

\paragraph{Pros} The local computation costs is $\mathcal{O}(d)$, which is same as FedAvg . It can perform well on heterogeneous and partial client FL. 

\paragraph{Cons} It is limited by its reliance on a diagonal approximation, which ignores off-diagonal elements that may capture important curvature details in highly non-convex settings. Need non-linear and twice differentiable objective function to approximate the Hessian diagonal. Additionally, due to the computation of local newton update, the local memory costs is slightly higher than that of FedAvg.

\subsubsection{\textbf{HWA}}

To address the issue of data heterogeneity across clients, which causes drift of local models compared to the global model, HWA proposes a novel aggregation method for the local models based on the diagonal elements of the Hessian corresponding to the model parameters of the output layer, represented as $\textbf{Z}^o_k$. In HWA, each client shares its locally trained model along with $\textbf{Z}^o_k$ to the server, and the server performs aggregation of the hidden layers of the model based on their isotropic weights, as used in FedAvg. For the aggregation of the output layers of the local models, it uses Eq. \ref{aggrega:Eq}, where $\beta_k = ||\textbf{Z}^o_k||_F + \epsilon$, with $\epsilon$ being a small constant that handles the issue of numerical instability, and $\textbf{w}^o_k$ is the output layer's parameters of the local model.

\begin{equation}
    \textbf{w}^o = \sum_{k=1}^{\mathcal{K}} \frac{\textbf{Z}^o_k}{\beta_k} \textbf{w}^o_k
    \label{aggrega:Eq}
\end{equation}

To compute the Hessian diagonal for output layers, it first approximates the Hessian corresponding to the output layer's model parameters using the square of the Jacobian matrix ($\widetilde{\textbf{H}}_k \equiv {\textbf{J} (f_k(\textbf{w}^o_k))}^T \textbf{J} (f_k(\textbf{w}^o_k)) \in \mathbb{R}^{r \times r}$, here $r < d$ is the size of output layer). The aggregation of the output layers of the local models, based on local Hessian information, takes into account the rate of convergence for each local model during the aggregation process, ensuring effective aggregation of the local models.

\paragraph{Empirical Analysis } Experiments on heterogeneously partitioned and partial client federated image classification on MNIST, FMNIST, FEMNIST and CIFAR10 datasets with multilayer perceptrons (MLPs)/CNN model demonstrate that the proposed method outperforms FedAvg, FedProx, FedCurve \cite{shoham2019overcoming}, and FedNL in terms of reducing training and test losses, as well as achieving better test accuracy in fewer communication rounds. The non-iid data partitions across clients are made by considering Dirichlet distribution using concentration parameter = 0.1.

\paragraph{Complexities} For approximating local Hessian diagonal and for computing local gradient, it requires overall $\mathcal{O}(d)$ computation cost and overall memory cost  $\mathcal{O}(r^2 + 2d)$ ($r<d$) in local clients. In server, aggregation of local models and local Hessian diagonals require overall $\mathcal{O}(\mathcal{K}r + \mathcal{K}(d-r))$ computation cost. For storing all the local models and output layer's Hessian diagonals, server needs overall $\mathcal{O}(\mathcal{K}r + \mathcal{K}d) $ memory cost. Server to client transmission costs is $\mathcal{O}(d)$ and client to server is $\mathcal{O}(d + r)$.

\paragraph{Pros} The local computation costs for calculation of Hessian is reduced from $\mathcal{O}(d^2)$ to $\mathcal{O}(r^2)$, where $r<d$ . It can perform well on heterogeneous  FL setting. 

\paragraph{Cons} It is limited by its reliance on a diagonal approximation, which ignores off-diagonal elements that may capture important curvature details in highly non-convex settings. Additionally, it increases $\mathcal{O}(r^2)$ computation load on local client as compared to FedAvg. Additionally, due to the computation of local newton update, the local memory costs is slightly higher than that of FedAvg.

\subsubsection{\textbf{Fed-Sophia}}
Fed-Sophia calculates the Hessian's diagonal using the Gauss-Newton-Bartlett (GNB) method \cite{liu2023sophia} and uses it to scale the local gradient while updating the local model using second-order optimization. Before computing the local Newton update, it computes the moving average of the current and previous local gradients as well as for Hessian's diagonal. During the computation of the Newton update, it applies a clipping operation to guard against inaccurate curvature information. Once the local models are trained, the server collects and aggregates them to update the global model.
 
\paragraph{Empirical Analysis } Experiments on heterogeneously partitioned federated image classification on MNIST and FMNIST datasets with multilayer perceptrons (MLPs)/CNN model demonstrate that the proposed method outperforms FedAvg and DONE in term of achieving better test accuracy in fewer communication rounds. The strategy of non-iid data partitions across clients is not mentioned in the paper.
  
\paragraph{Complexities} For computing local gradient and approximating local Hessian diagonal, it requires overall $\mathcal{O}(d)$ computation cost and overall memory costs  $\mathcal{O}(5d)$ in local clients for storing local model, current local gradient, previous local gradient, current local Hessian diagonal and previous local Hessian diagonal. In server, aggregation of local models requires overall $\mathcal{O}(\mathcal{K}d)$ computation costs. For storing all the local models needs overall $\mathcal{O}(\mathcal{K}d) $ memory. Server to client transmission costs is $\mathcal{O}(d)$ and client to server is $\mathcal{O}(d)$, which is same as FedAvg.

\paragraph{Pros} The local computation costs for calculation of Hessian is reduced from $\mathcal{O}(d^2)$ to $\mathcal{O}(d)$ . It can perform well on heterogeneous  FL setting. It has same transmission costs as FedAvg. 

\paragraph{Cons} It is limited by its reliance on a diagonal approximation, which ignores off-diagonal elements that may capture important curvature details in highly non-convex settings. Additionally, it increases computation time while updating the local model for using GNB on local client as compared to FedAvg. Additionally, due to the computation of local newton update, the local memory costs is slightly higher than that of FedAvg.

\subsection{\textbf{Nyström method-based Hessian approximation}}
Methods including FONN  \cite{nagarajufonn_23} and Nys-FL \cite{mrinmaynys-fl_23} use Nys-Newton \cite{singh2021nys} for optimizing global model in FL. The key advantage of using Nys-Newton in FL is that , it approximate the Hessian by performing Nyström method on randomly sketching $r \ll d$ columns of  the exact Hessian and directly compute Newton update without storing full Hessian matrix.  

\subsubsection{\textbf{FONN}}
Working procedure of FONN is similar to that of GIANT, only modification is the replacement of CG-based local update calculation with the Nys-Newton \cite{singh2021nys}. It calculates local Newton update by performing Nys-Newton with global gradient. As it uses Nys-Newton, the convergence rate of FONN is linear-quadratic for non-convex objective. 

\paragraph{Empirical Analysis } The performance of FONN is validated through experiments on heterogeneously (Dirichlet 0.2) partitioned MNIST, FMNIST and SVHN datasets, where $20\%$ data is used for checking the performance of the global model. The federated classification tasks for each of these datasets is performed using logistic regression. The outcomes from the experiments shows that FONN outperforms SCAFFOLD, GIANT and DONE in terms of reducing train and test losses as well as achieving better test accuracy in lower communication rounds. \\

\paragraph{Complexities}  
As FONN involves the computation of the global gradient and global Newton update separately, it requires four transmissions of $\mathcal{O}(d)$ from the server to the client and from the client to the server. To compute local Newton update using randomly sketched $r$ columns of the Hessian, it requires overall $\mathcal{O}(d)$ computation costs and $\mathcal{O}(rd) + 2d$ memory costs for storing $r$ columns of Hessian along with the gradient and local model. The overall server computation cost of FONN is $\mathcal{O}(\mathcal{K}d)$ for averaging local models, and the overall memory costs is $\mathcal{O}(\mathcal{K}d)$ for local models.

\paragraph{Pros} FONN is advantageous in terms of its ability to handle non-convex objective (as Nys-Newton is applicable to that case) and its linear-quadratic convergence.

\paragraph{Cons} FONN is designed for a twice-differentiable objective function. It also requires increased time to complete one communication round due to the four transmissions. Additionally, due to the computation of Hessian columns, the local memory costs is slightly higher ( $r \ll d$ times more) than that of FedAvg.

\subsubsection{\textbf{Nys-FL}}

Nys-FL performs the Nys-Newton method \cite{singh2021nys} with the global gradient and generated server data while updating the global model. To generate data, it computes global Gaussian mixture models by aggregating locally trained Gaussian mixture models. Nys-FL updates the local model using the SGD optimizer and shares it with the server. The server computes the gradient for each client from the SGD update rule and then aggregates all the local gradients to compute the global gradient. The use of server-based Nys-Newton leads to linear-quadratic convergence of the global model with a slight increase in the server computation and memory costs compared to FedAvg.

\paragraph{Empirical Analysis } The performance of Nys-FL is validated through experiments on heterogeneously (Dirichlet 0.2) partitioned MNIST, FMNIST and SVHN datasets, where $20\%$ data is used for checking the performance of the global model. The federated classification tasks for each of these datasets is performed using logistic regression. The outcomes from the experiments shows that Nys-FL outperforms FedProx, SCAFFOLD, GIANT and DONE in terms of reducing train and test losses as well as achieving better test accuracy in lower communication rounds. 

\paragraph{Complexities}  
The local computation and memory costs of Nys-FL are similar to those of FedAvg, i.e., $\mathcal{O}(d)$ and $\mathcal{O}(2d)$, respectively ($\mathcal{O}(d)$ for the gradient and $\mathcal{O}(d)$ for the local model). On the server, the aggregation of local gradients and local models is associated with overall computation cost of $\mathcal{O}(\mathcal{K}d)$ and memory cost of $\mathcal{O}(2\mathcal{K}d + d)$ ($\mathcal{O}(d)$ for storing the previous global model). Updating the global model using Nys-Newton requires $\mathcal{O}(d)$ computation costs and $\mathcal{O}(rd + 2d)$ memory costs. Therefore, the overall server computation costs of Nys-FL are $\mathcal{O}(\mathcal{K}d)$. The overall server memory costs of Nys-FL are $\mathcal{O}(\mathcal{K}d)$ and $\mathcal{O}(2\mathcal{K}d + d + n_s)$ if $r < \mathcal{K}$, and $\mathcal{O}(rd + 2d + n_s)$ otherwise ($n_s$ is number of generated samples). The transmission costs of Nys-FL are the same as FedAvg in both directions, which is one of the key advantages of Nys-FL.

\paragraph{Pros} Nys-FL is advantageous in terms of its ability to handle non-convex objective (as Nys-Newton is applicable to that case), its linear-quadratic convergence.

\paragraph{Cons} Nys-FL is designed for a twice-differentiable objective function. Additionally, due to use of Nys-Newton , the server memory costs is higher than that of FedAvg. Due to use of local GMMs, it violates local data privacy. 

\subsection{\textbf{One-rank approximation of Hessian}}

One rank approximation of Hessian involves approximating it with the outer product of two vectors  ($\textbf{u}, \textbf{v} \in \mathbb{R}^{d}$ as shown in Eq. \ref{oneRHAprox:eq}

\begin{equation}
    \widetilde{\textbf{H}} = \textbf{u}\textbf{v}^T
    \label{oneRHAprox:eq}
\end{equation}
Existing methods in this category involves FReNG \cite{senfreng_23}, FAGH \cite{sen2024fagh} etc.

\subsubsection{\textbf{FReNG}}
FReNG uses a regularized Fisher Information Matrix (FIM) $\textbf{F}^t$ built using the global gradient ($\textbf{g}^t$), as given in Eq. \ref{Eq:21} (here, $\rho \textbf{I}$ is the regularization term added to overcome the issue of numerical instability when inverting the FIM)
), and applies the Sherman-Morrison formula of matrix inversion to Eq. \ref{Eq:22} to compute the global Newton update $\widetilde{\Delta^t}$ directly. This is a key advantage of FReNG, as it does not require storing or calculating the full FIM. The main motivation for replacing the Hessian with the FIM is that, for probabilistic objectives, the Hessian is equivalent to the FIM. So, FReNG is limited to the applications with probabilistic objectives.
\begin{equation}
    \textbf{F}^t = \textbf{g}^t {\textbf{g}^t}^T + \rho \textbf{I}
    \label{Eq:21}
\end{equation}

\begin{equation}
    \Delta^t = {(\textbf{g}^t {\textbf{g}^t}^T + \rho \textbf{I})}^{-1} \textbf{g}^t
    \label{Eq:22}
\end{equation}

FReNG computes the global gradient by aggregating local gradients. Instead of directly using this global gradient, FReNG computes the first moment of this gradient and uses it when computing the Newton update.

\paragraph{Empirical Analysis } The performance of FReNG is validated through experiments on heterogeneously (Dirichlet 0.2) partitioned MNIST, FMNIST and CIFAR10 datasets, where $20\%$ data is used for checking the performance of the global model. The federated classification tasks for each of these datasets is performed using logistic regression. The outcomes from the experiments shows that FReNG outperforms FedProx, SCAFFOLD, FedInit \cite{sun2023understanding}, FedGA \cite{dandi2022implicit} and DONE in terms of reducing train and test losses as well as achieving better test accuracy in lower communication rounds.

\paragraph{Complexities}  
The local computation and memory cost of FReNG are similar to those of FedAvg, i.e., $\mathcal{O}(d)$ and $\mathcal{O}(2d)$, respectively ($\mathcal{O}(d)$ for the gradient and $\mathcal{O}(d)$ for the local model). On the server, the aggregation of local gradients, computing the momentum of global gradient and updating the global model using regularized FIM, FReNG requires overall $\mathcal{O}(\mathcal{K}d)$ computation cost and $\mathcal{O}(\mathcal{K}d) + 2d$ memory cost (for storing local gradients, previous global gradient and global model). The transmission cost of FReNG is same as FedAvg in both directions, which is one of the key advantages of FReNG.

\paragraph{Pros} FReNG effectively utilizes second-order optimization in FL with the similar local computation costs of FedAvg. The transmission costs associated with Hessian is reduced from $\mathcal{O}(d^2)$ to  $\mathcal{O}(d)$. The server computation is also nearly same as FedAvg. 

\paragraph{Cons} FreNG is applicable for probabilistic objective functions only.

\subsubsection{\textbf{FAGH}} 
In the server, FAGH aggregates all the clients' Hessian first rows to compute the global Hessian's first row ($\textbf{v}^t$) and aggregates all the local gradients to compute the global gradient ($\textbf{g}^t$). FAGH uses the first row of the global Hessian to approximate the global Hessian using Eq. \ref{Eq:23}, where $\textbf{z}^t = \frac{\textbf{v}^t}{\textbf{v}^t[0]}$, and $\textbf{v}^t[0]$ is the first element of $\textbf{v}^t$. 

\begin{equation}
    \widetilde{\textbf{H}}^t = \textbf{z}^t {\textbf{v}^t}^T 
    \label{Eq:23}
\end{equation}

Like FReNG, FAGH directly calculates the global model update ($\widetilde{\Delta^t}$) using the Sherman-Morrison formula for matrix inversion in Eq. \ref{Eq:24}, where $\rho \textbf{I}$ is the regularization term, and uses it to update the global model. Before applying the Sherman-Morrison formula, FAGH computes the first moments of the first row of the global Hessian and the global gradient.

\begin{equation}
    \Delta^t = \left( \textbf{z}^t {\textbf{v}^t}^T + \rho \textbf{I} \right)^{-1} \textbf{g}^t
    \label{Eq:24}
\end{equation}

\paragraph{Empirical Analysis } The performance of FAGH is validated through experiments on heterogeneously partitioned FMNIST, EMNIST and CIFAR10 datasets, where $20\%$ data is used for checking the performance of the global model. For CIFAR10 and FMNIST, Dirichlet distribution with concentration parameter = 0.2 is used for creating non-iid data across client and for EMNIST the data is equally divided across clients considering class labels. The federated classification tasks for these datasets is performed using logistic regression for EMNIST, CNN model for FMNIST and LeNet5 model for CIFAR10. The outcomes from the experiments shows that FAGH outperforms SCAFFOLD, FedExP \cite{jhunjhunwala2023fedexp}, FedGA \cite{dandi2022implicit}, GIANT and DONE in terms of reducing train and test losses as well as achieving better test accuracy in lower time and communication rounds.

\paragraph{Complexities}  
For computing first row of Hessian and gradient, the local computation costs of FAGH is  $\mathcal{O}(d)$ and memory costs of $\mathcal{O}(3d)$ for storing gradient, first row of Hessian and local model. For aggregating local first-rows,  aggregating local gradients and updating the global model using Sherman-Morrison formula, server requires overall $\mathcal{O}(\mathcal{K}d)$ computation costs. For storing local Hessian's first rows, local gradients, previous global gradient, previous global Hessian's first row and previous global model, server needs overall $\mathcal{O}(2\mathcal{K}d) + 3d$ memory costs. Server to client transmission costs of FAGH is $\mathcal{O}(d)$ where client to server transmission costs is $\mathcal{O}(2d)$

\paragraph{Pros} As FAGH computes the exact Hessian approximation using Eq. \ref{Eq:23} and directly computes the model update using the Sherman-Morrison formula, it can achieve linear-quadratic convergence of the global model depending on the choice of the Hessian regularization parameter (for convex function). It can reduce Hessian transmission cost from $\mathcal{O}(d^2)$ to $\mathcal{O}(d)$.

\paragraph{Cons} Application of Eq. \ref{Eq:23} requires that the objective function is twice differentiable and non-linear. Due to sending the first row of the Hessian, it slightly increases client-to-server transmission costs as compared to FedAvg. 
%%%%%%%%%%%%%%%%%%%%%%%%%%%%%%%%%%%%%%%%%%%%%%%%%%%%%%%%

\begin{table*}
    \centering
    \caption{Complexities of different second-order FL methods in terms of local and server memory, computation costs, and server-client/client-server transmission costs, here $r , r' < d$ are integer valued number and $n_s$ is the number of samples generated in the server. The computation cost for the local client is considered per sample.}
    \renewcommand{\arraystretch}{1.5} % Adjust row spacing
    {\begin{tabular}{|>{\centering\arraybackslash}p{2cm}
                |>{\centering\arraybackslash}p{1.5cm}
                |>{\centering\arraybackslash}p{2cm}
                |>{\centering\arraybackslash}p{2cm}
                |>{\centering\arraybackslash}p{2cm}
                |>{\centering\arraybackslash}p{2cm}
                |>{\centering\arraybackslash}p{2cm}|}
        \hline
        \textbf{Methods} & \textbf{Local Computation Cost} & \textbf{Server Computation Cost} & \textbf{Local Memory Cost} & \textbf{Server Memory Cost} & \textbf{Transmission Cost (Client$\rightarrow$ Server)} & \textbf{Transmission Cost (Server$\rightarrow$ Client)}\\ 
        \hline \hline
        \textbf{DANE [2014]}&
        $\mathcal{O}(d)$ & 
        $\mathcal{O}(K d)$ & 
        $\mathcal{O}(4d)$ & 
        $\mathcal{O}(K d + d)$ & 
        $\mathcal{O}(d)$ & 
        $\mathcal{O}(d)$\\
        \hline
        \textbf{AIDE [2016]}&
        $\mathcal{O}(d)$ & 
        $\mathcal{O}(K d)$ & 
        $\mathcal{O}(4d)$ & 
        $\mathcal{O}(K d + d)$ & 
        $\mathcal{O}(d)$ & 
        $\mathcal{O}(d)$\\
        \hline 
        \textbf{FedDANE [2019]}&
        $\mathcal{O}(d)$ & 
        $\mathcal{O}(K d)$ & 
        $\mathcal{O}(4d)$ & 
        $\mathcal{O}(K d + d)$ & 
        $\mathcal{O}(d)$ & 
        $\mathcal{O}(d)$\\
        \hline 
        \textbf{DiSCO [2015]}&
        $\mathcal{O}(d)$ & 
        $\mathcal{O}(K d)$ & 
        $\mathcal{O}(3d)$ & 
        $\mathcal{O}(K d + d)$ & 
        $\mathcal{O}(d)$ & 
        $\mathcal{O}(2d)$\\
        \hline 
        \textbf{GIANT [2018]}&
        $\mathcal{O}(d)$ & 
        $\mathcal{O}(K d)$ & 
        $\mathcal{O}(4d)$ & 
        $\mathcal{O}(K d + d)$ & 
        $\mathcal{O}(d)$ & 
        $\mathcal{O}(d)$\\
        \hline 
        \textbf{DONE [2022]}&
        $\mathcal{O}(d)$ & 
        $\mathcal{O}(K d)$ & 
        $\mathcal{O}(4d)$ & 
        $\mathcal{O}(K d + d)$ & 
        $\mathcal{O}(d)$ & 
        $\mathcal{O}(d)$\\
        \hline 
        \textbf{Local Newton [2021]}&
        $\mathcal{O}(d)$ & 
        $\mathcal{O}(K d)$ & 
        $\mathcal{O}(3d)$ & 
        $\mathcal{O}(K d)$ & 
        $\mathcal{O}(d)$ & 
        $\mathcal{O}(d)$\\
        \hline  \hline
        \textbf{FedNL [2022]}&
        $\mathcal{O}(d^2)$ &
        $\mathcal{O}(d^3)$ & 
        $\mathcal{O}(d^2 + 2d)$ &
        $\mathcal{O}(d^2 + 2d)$ &
        $\mathcal{O}(2d)$ & 
        $\mathcal{O}(d)$\\
        \hline 
        \textbf{Basis Matters [2022]}&
        $\mathcal{O}(d^2)$ &
        $\mathcal{O}(d^3)$ & 
        $\mathcal{O}(d^2 + 2d)$ &
        $\mathcal{O}(d^2 + 2d)$ &
        $\mathcal{O}(r'^{2} + d)$
        $(r' \ll d)$& 
        $\mathcal{O}(d)$\\
        \hline 
        \textbf{FedNew [2022]}&
        $\mathcal{O}(d^3)$ & 
        $\mathcal{O}(K d)$ & 
        $\mathcal{O}(d^2 + 2d)$ &
        $\mathcal{O}(K d + d)$ & 
        $\mathcal{O}(r)$
        $(r < d)$&
        $\mathcal{O}(d)$\\
        \hline 
        \textbf{SHED [2024]}&
        $\mathcal{O}(d^2)$ & 
        $\mathcal{O}(K d^2)$ & 
        $\mathcal{O}(d^2 + 2d)$ & 
        $\mathcal{O}(K d^2+ K\cdot d + d)$ & 
        $\mathcal{O}(r' d + d)$& 
        $\mathcal{O}(d)$\\
        \hline
        \textbf{FedNS [2024]}&
        $\mathcal{O}(d^2)$ &
        $\mathcal{O}(K d^2)$ & 
        $\mathcal{O}(d^2 + 2d)$ &  
        $\mathcal{O}(K d^2+ K\cdot d + d)$ & 
        $\mathcal{O}(rd + d)$
        $(r \ll d)$& 
        $\mathcal{O}(d)$\\
        \hline \hline
        \textbf{FedSSO [2022]}&
        $\mathcal{O}(d)$ & 
        $\mathcal{O}(d^2)$ & 
        $\mathcal{O}(2d)$ & 
        $\mathcal{O}(d^2 + 4d)$ & 
        $\mathcal{O}(d)$ & 
        $\mathcal{O}(d)$\\
        \hline 
        \textbf{DQN-Fed [2025]}&
        $\mathcal{O}(d^2)$ & 
        $\mathcal{O}(K d)$ & 
        $\mathcal{O}(d^2 + 4d)$ &
        $\mathcal{O}(2K d + d)$ & 
        $\mathcal{O}(d)$ & 
        $\mathcal{O}(d)$\\
        \hline \hline
        \textbf{FopLAHD [2023]}&
        $\mathcal{O}(d)$ & 
        $\mathcal{O}(K d)$ & 
        $\mathcal{O}(4d)$ & 
        $\mathcal{O}(K d + 3d)$ & 
        $\mathcal{O}(2d)$ & 
        $\mathcal{O}(d)$\\
        \hline 
        \textbf{LTDA [2023]}&
        $\mathcal{O}(d)$ & 
        $\mathcal{O}(K d)$ & 
        $\mathcal{O}(3d)$ & 
        $\mathcal{O}(K d)$ & 
        $\mathcal{O}(d)$ & 
        $\mathcal{O}(d)$\\
        \hline 
        \textbf{HWA [2023]}&
        $\mathcal{O}(d)$ & 
        $\mathcal{O}(K r + K (d-r))$ & 
       $\mathcal{O}(r^2 + 2d)$ &
        $\mathcal{O}(K r + K d)$ & 
        $\mathcal{O}(d + r)$ & 
        $\mathcal{O}(d)$\\
        \hline
        \textbf{Fed-Sophia [2024]}&
        $\mathcal{O}(d)$ & 
        $\mathcal{O}(K d)$ & 
        $\mathcal{O}(5d)$ & 
        $\mathcal{O}(K d)$ & 
        $\mathcal{O}(d)$ & 
        $\mathcal{O}(d)$\\
        \hline \hline
        \textbf{FONN [2023]}&
        $\mathcal{O}(d)$ & 
        $\mathcal{O}(K d)$ & 
        $\mathcal{O}(rd+2d)$ & 
        $\mathcal{O}(K d)$ & 
        $\mathcal{O}(d)$ & 
        $\mathcal{O}(d)$\\
        \hline 
        \textbf{Nys-FL [2023]}&
        $\mathcal{O}(d)$ & 
        $\mathcal{O}(K d)$ & 
        $\mathcal{O}(2d)$ & 
        $\mathcal{O}(2K d + d + n_s)$& 
        $\mathcal{O}(d)$ & 
        $\mathcal{O}(d)$\\
        \hline \hline
    
        \textbf{FReNG [2023]}&
        $\mathcal{O}(d)$ & 
        $\mathcal{O}(K d)$ & 
        $\mathcal{O}(2d)$ & 
        $\mathcal{O}(K d +2d)$ & 
        $\mathcal{O}(d)$ & 
        $\mathcal{O}(d)$\\
        \hline 
        \textbf{FAGH [2024]}&
        $\mathcal{O}(d)$ & 
        $\mathcal{O}(K d)$ & 
        $\mathcal{O}(3d)$ & 
        $\mathcal{O}(2K d+3d)$ & 
        $\mathcal{O}(2d)$ & 
        $\mathcal{O}(d)$\\
        \hline 
    
    \end{tabular}\\

    \label{tab:my_label1}}
\end{table*}
%%%%%%%%%%%%%%%%%%%%%%%%%%%%%%%%%%%%%%%%%%%%%%%%%%%%%%%%%%
\begin{table*}
    \centering
    \caption{Descriptions of different second-order FL methods with their contributions, applications, transmission overhead, computation of newton update and their convergence property, here FCP and PCP mean full client participation and partial client participation respectively and transmission number denotes number of transmissions required to complete one communication round }
    \renewcommand{\arraystretch}{1.5} % Adjust row spacing
    {\begin{tabular}{|>{\centering\arraybackslash}p{1.2cm}
                |>{\arraybackslash}p{6cm}
                |>{\centering\arraybackslash}p{0.5cm}
                |>{\centering\arraybackslash}p{0.5cm}
                |>{\centering\arraybackslash}p{0.9cm}
                |>{\centering\arraybackslash}p{1.0cm}
                |>{\centering\arraybackslash}p{2cm}
                |>{\centering\arraybackslash}p{2.5cm}|}
        \hline
        \textbf{Methods} & \centering\textbf{Contribution} & \textbf{iid / non-iid} & \textbf{FCP/ PCP} & \textbf{Newton Update} & \textbf{ Transmission number} & \textbf{Objective} & \textbf{Convergence}\\ 
        \hline \hline
        \textbf{DANE}&
        Modifies local objectives with two regularization terms &
        iid & 
        FCP & 
        local & 
        4 & 
        convex&
        linear-quadratic
        \\
        \hline 
        \textbf{AIDE}&
        Enhances DANE by solving local objectives inexactly with algorithms like Quartz, then integrates catalyst& 
        non-iid & 
        FCP & 
        local & 
        4 & 
        convex&
        linear\\
        \hline 
        \textbf{FedDANE}&
        Shows the issues of non-iid data and PCP in DANE & 
        non-iid & 
        PCP & 
        local & 
        4 & 
        convex&
        ----\\
        \hline 
        \textbf{DiSCO}&
        Utilizes preconditioned CG method while solving local subproblem& 
        iid & 
        FCP & 
        Local & 
        4& 
        convex&
        super-linear (self-concordant)\\
        \hline 
        \textbf{GIANT}&
        Utilizes CG method for solving local sub-problems with the global gradient and computes harmonic mean of local updates& 
        iid & 
        FCP & 
        local & 
        4& 
        convex&
        linear-quadratic\\
        \hline \textbf{DONE}&
        Utilizes Richardson iteration to compute local update& 
        non-iid & 
        FCP & 
        local & 
        4& 
        convex &
        linear-quadratic\\
        \hline \textbf{Local Newton}&
         Solves local sub-problem using the CG method with the local gradient and increased local iterations ($r$) &
        iid & 
        FCP & 
        local & 
        2 & 
        convex&
        linear for $r=1$ sub-linear for $r>1$  \\
        \hline \hline \textbf{FedNL}&
        Compresses the Hessian before sending it to the server& 
        non-iid & 
        FCP + PCP& 
        server & 
        2 & 
        non-convex&
        super-linear (distance to optima)\\
        \hline\textbf{Basis Matters}&
        Compresses the Hessian before sending it to the server& 
        non-iid & 
        FCP & 
        server & 
        2 & 
        non-convex&
        super-linear\\
        \hline\textbf{FedNew}&
        Computes local Newton update using one ADMM iteration & 
        iid & 
        FCP & 
        local & 
        2 & 
        convex& ----
        \\
        \hline\textbf{SHED}&
        Computes local eigen vectors in each client and uses it to approximate the local Hessian in the server & 
        non-iid & 
        FCP & 
        server & 
        2 & 
        convex + non-convex&
        super-linear (asymptotic)\\
        \hline \textbf{FedNS}& Sketches the local Hessian before sharing to the server
        & 
        non-iid & 
        FCP & 
        server & 
        2& 
        Convex&
        super-linear\\
        \hline \hline \textbf{FedSSO}&
        Approximates the global Hessian using the BFGS algorithm in the server with the global gradient & 
        non-iid & 
        FCP & 
        server & 
        2 & 
        convex + non-convex&
        sub-linear (convex)\\
        \hline \textbf{DQN-Fed}&
        Orthogonalizes local gradients and computes the aggregation weights based on Hessian curvature computed through BFGS algorithm & 
        non-iid & 
        PCP & 
        local & 
        2 & 
        convex + non-convex&
        linear-quadratic (convex)\\
        \hline \hline \textbf{FopLAHD}&
        Approximates Hessian diagonal in local client and makes the global Hessian diagonal free from noise& 
        non-iid & 
        PCP & 
        server & 
        2& 
        convex + non-convex& ---\\
        \hline  \textbf{LTDA}&
        Linear time approximation of Hessian diagonal in the local client using first row of the Hessian & 
        non-iid & 
        PCP & 
        local & 
        2& 
        convex + non Convex&----
        \\
        \hline \textbf{HWA}&
        Aggregates local models based on local Hessian diagonals& 
        non-iid & 
        PCP & 
        local & 
        2 & 
        convex + non-convex&---
        \\
        \hline \textbf{Fed- Sophia}&
        Computes local Hessian diagonal using GNB method& 
        non-iid & 
        FCP & 
        local & 
        2 & 
        convex + non-convex&---
       \\
        \hline \hline \textbf{FONN}&
        Uses Nys-Newton with the global gradient while updating local model & non-iid &
        PCP & 
        local & 
        4& 
        convex + non-convex& linear-quadratic\\
        \hline \textbf{Nys-FL}&
        Uses Nys-Newton in the server with the GMMs-based generated data & non-iid &
        PCP & 
        server & 
        2& 
        convex + non-convex& linear-quadratic\\

        \hline \hline \textbf{FReNG}&
        Uses regularized FIM while updating the global model & non-iid &
        PCP & 
        server & 
        2& 
        convex + non-convex& linear-quadratic (convex)\\

        \hline \textbf{FAGH}&
        Approximates global Hessian in the server using its first row & non-iid&
        PCP & 
        server & 
        2& 
        convex + non-convex& linear-quadratic (convex)\\ \hline

    \end{tabular}\\

    \label{tab:my_label2}}
\end{table*}

%%%%%%%%%%%%%%%%%%%%%%%%%%%%%%%%%%%%%%%%%%%%%%%%%%%%%%

%%%%%%%%%%%%%%%%%%%%%%%%%%%%%%%%%%%%%%%%%%%%%%%%%%%%%%%

\begin{table*}
    \centering
    \caption{Experiments and results of different methods as presented in the respective papers, here FPC : full client participation and PCP : partial client participation.}
    \renewcommand{\arraystretch}{1} % Adjust row spacing
    {\begin{tabular}{|>{\arraybackslash}p{1cm}
                |>{\arraybackslash}p{2.7cm}
                |>{\arraybackslash}p{3cm}
                |>{\arraybackslash}p{1cm}
                |>{\arraybackslash}p{0.5cm}
                |>{\arraybackslash}p{2.5cm}
                |>{\arraybackslash}p{4cm}
                |>{\arraybackslash}p{3cm}|}
        \hline
        \textbf{Methods} & \textbf{Data} & \textbf{Model/Loss Function} & \textbf{Par-tition} & \textbf{FCP/ PCP} & \textbf{Compared Method} & \textbf{Results}\\ 
        \hline \hline
        \textbf{DANE}&
        COV1, ASTRO-PH, MNIST & 
        Regression with regularized hinge loss& 
        iid & 
        FCP & 
        One-short parameter average, ADMM & 
        Faster reduction in training loss\\
        \hline 
        \textbf{AIDE}&
        RCv1, CovType, Real-Sim, URL& 
        Logistic regression using hinge loss& 
        iid & 
        FCP & 
        CoCoA+, DANE & 
        Faster reduction of function suboptimality\\
        \hline 
        \textbf{FedDANE}&
        FEMNIST, Shakespeare, Sent-140  & 
        Convex model for FEMNIST, DDN model for others & 
        non-iid & 
        PCP & 
        FedAvg, FedProx& 
        Under performs in terms of reducing training loss\\
        \hline 
        \textbf{DiSCO}&
        LIBSVM & 
        Binary classification using logistic regression & 
        iid & 
        FCP & 
        ADMM, AFG, L-BFGS, DANE& 
        Faster reduction in training loss\\
        \hline 
        \textbf{GIANT}&
        MNIST8M, Covtype & 
        Classification using logistic regression & 
        iid & 
        FCP & 
        AGD, L-BFGS, DANE & 
        faster reduction in classification error\\
        \hline 
        \textbf{DONE}&
        MNIST, FEMNIST, HAR & 
        Classification using Logistic Regression  & 
        non-iid & 
        FCP & 
        FedAvg, GIANT, FEDL, DANE & 
        Faster reduction in training loss and improving test accuracy\\
        \hline 
        \textbf{Local Newton}&
        LIBSVM & 
        Classification using Logistic Regression & 
        iid & 
        FCP & 
        GIANT, L-BFGS, Local SGD& 
        Faster reduction in training loss and improving test accuracy\\
        \hline \hline
        \textbf{FedNL}&
        LIBSVM& 
        Binary classification using regularized Logistic Regression & 
        iid & 
        FCP & 
        DIANA, ADIANA, shifted Local GD, DINGO & 
        Faster reduction of classification error\\
        \hline 
        \textbf{Basis Matters}&
        LIBSVM& 
        Binary classification using regularized Logistic Regression& 
        iid & 
        FCP & 
        Vanilla-GD, DIANA, ADIANA, FedNL & 
        Faster reduction of classification error\\
        \hline 
        \textbf{FedNEW}&
        LIBSVM& 
        Binary classification using regularized Logistic Regression& 
        iid & 
        FCP & 
        FedAvg, FedNL& 
        Faster reduction of classification error\\
        \hline
        \textbf{SHED}&
        Million song, FMNIST, EMNIST, w8a& 
        Logistic Regression& 
        non-iid + iid & 
        PCP & 
        BFGS, GIANT, FedNL& 
        Faster reduction of classification error\\
        \hline 
        \textbf{FedNS}&
        LIBSVM& 
        Classification using Logistic Regression& 
        iid & 
        FCP & 
        FedAvg, FedNewton, FedNew, FedNL& 
        Faster reduction of classification error\\
        \hline \hline
        \textbf{FedSSO}&
        MNIST, EMNIST, CIFAR10, LIBSVM& 
        Logistic Regression + CNN& 
        non-iid & 
        FCP & 
        FedSGD, FedAvg, FedAC, FedOpt, SCAFFOLD, FedDANE, FedNL& 
        Faster reduction of training loss and improving test accuracy\\
        \hline 
        \textbf{DQN-Fed}&
        CIFAR10, CIFAR100, Shakeseare, FMNIST, TinyImagenet, CINIC10& 
        DNN + CNN with categorical cross entropy& 
        non-iid & 
        PCP & 
        qFFL, TERM, FedMGDA, Ditto, FedLF, FedHEAL, FedNL, FedNew& 
        Faster improving test accuracy\\
        \hline \hline
        \textbf{Fop- LADH}&
        MNIST, FMNIST& 
        multi logistic regression using categorical cross entropy loss& 
        non-iid & 
        PCP & 
        FedProx, SCAFFOLD, DONE, GIANT& 
        Faster reduction of training and test losses and improving test accuracy\\
        \hline 
        \textbf{LTDA}&
        MNIST, FMNIST& 
        multi logistic regression using categorical cross entropy loss& 
        non-iid & 
        PCP & 
        FedProx, SCAFFOLD, DONE& 
        Faster reduction of training and test losses and improving test accuracy\\
        \hline 

        \textbf{HWA}&
        MNIST, FMNIST, FEMNIST, CIFAR10& 
        MLP and CNN using categorical cross entropy loss& 
        non-iid & 
        FCP & 
        FedAvg, FedProx, FedCurve, FedNL& 
        Faster reduction of training and test losses and improving test accuracy\\
        \hline 
        \textbf{Fed- Sophia}&
        MNIST, FMNIST& 
        MLPs and CNN using categorical cross entropy loss& 
        non-iid & 
        FCP & 
        FedAvg, DONE& 
        Faster improving test accuracy\\
        \hline \hline
        \textbf{FONN}&
        MNIST, FMNIST, SVHN& 
        multi logistic regression using categorical cross entropy loss& 
        non-iid & 
        PCP & 
        SCAFFOLD, DONE, GIANT& 
        Faster reduction of training and test losses and improving test accuracy\\
        \hline 
        \textbf{Nys-FL}&
        MNIST, FMNIST, SVHN& 
        multi logistic regression using categorical cross entropy loss& 
        non-iid & 
        PCP & 
        FedProx, SCAFFOLD, DONE, GIANT& 
        Faster reduction of training and test losses and improving test accuracy\\
        \hline \hline
        \textbf{FReNG}&
        MNIST, FMNIST, CIFAR10& 
        multi logistic regression using categorical cross entropy loss& 
        non-iid & 
        PCP & 
        FedProx, SCAFFOLD, FedInit, FedGA, DONE, GIANT& 
        Faster reduction of training and test losses and improving test accuracy\\
        \hline 
        \textbf{FAGH}&
        EMNIST, FMNIST, CIFAR10& 
        multi logistic regression and CNN using categorical cross entropy loss& 
        non-iid & 
        PCP & 
        SCAFFOLD, FedExP, FedGA, DONE, GIANT& 
        Faster reduction of training and test losses and improving test accuracy\\
        \hline 

    \end{tabular}\\
    \label{tab:my_label3}}
\end{table*}

%%%%%%%%%%%%%%%%%%%%%%%%%%%%%%%%%%%%%%%%%%%%%%%%%%%%%%%

\section{Comparative Evaluations}

In this section, we provide theoretical comparisons among different methods across various categories, including their complexities, main contributions, applicability, and convergence. We also present empirical comparisons based on datasets, models, loss functions, FL partitioning, compared methods, and results. The empirical comparisons are made by using experimental information from the respective papers of the methods.

\subsection{Theoretical Comparisons}
The theoretical comparisons of different methods are done through Tables \ref{tab:my_label1} and \ref{tab:my_label2}.  

From Table \ref{tab:my_label1}, we can see that the overall local computation and memory costs of all categories, except for those based on full local Hessian computation and Quasi-Newton methods, are either the same as or comparable to FedAvg. In contrast, the methods under the other two categories (except FedSSO, which has the same local computation and memory costs as FedAvg) incur quadratic computation and memory complexities, which could become a major bottleneck when applying these methods on resource-constrained local devices. The server computation and memory costs for most methods in the full local Hessian computation and Quasi-Newton categories are very high (except for FedNew and DQN-Fed, which have comparable requirements to FedAvg), necessitating a central server with sufficient computational resources. Server complexities for methods in other categories are comparable to those of FedAvg. The server-to-client transmission cost for all methods (except Disco) is $\mathcal{O}(d)$, similar to FedAvg, whereas Disco incurs a $\mathcal{O}(2d)$ cost. We notice that the transmission costs for the majority of methods are nearly identical to those of FedAvg, except for FedNL, FAGH, and FopLADH, which have $\mathcal{O}(2d)$ cost.

Through Table \ref{tab:my_label2}, we describe the major contributions and convergence properties of each method, along with the conditions applied during the design of each method, such as whether the data is IID or non-IID across clients, whether full client participation (FCP) or partial client participation (PCP) is used, and whether the objective is convex or non-convex. Additionally, we show the location of the computation of the Newton update, which can help in selecting an algorithm based on the resource availability on the server or clients. We also display the number of transmissions required by each method to complete one communication round. From Table \ref{tab:my_label2}, we observe that most of the Hessian-free FL methods are designed for convex objectives and IID data distribution across clients. Furthermore, these Hessian-free methods require increased time due to four transmissions per communication round. In contrast, most methods in other categories are applicable to non-IID data and both convex and non-convex objectives. Similar to FedAvg, these methods have the same number of transmissions as FedAvg, i.e., 2 (except for FONN, which has 4 transmissions, like the Hessian-free methods). From Table \ref{tab:my_label2}, we also observe that most methods in the diagonal approximation, Nyström method-based, and one-rank approximation categories are designed to allow partial client participation (PCP), while most methods in other categories assume full client participation (FCP). We note that methods in the Hessian-free category exhibit linear or linear-quadratic convergence, methods with local Hessian computation can achieve super-linear convergence, quasi-Newton methods can achieve sub-linear or linear-quadratic convergence, and methods in the Nyström method-based and one-rank approximation categories can achieve linear-quadratic convergence. All convergence properties are provided for convex objectives. 

The summarized findings from these two tables (Tables \ref{tab:my_label1} and \ref{tab:my_label2}) are presented in Table \ref{tab:my_label4}, which provides an overview of the methods in each category.

\subsection{Empirical Comparisons}
Table \ref{tab:my_label3} presents the empirical experiments and results conducted in the respective papers for each method. From this table, it can be observed that most of the methods in the Hessian-free category and full local Hessian computation are validated using IID data distribution and full client participation. In contrast, the majority of methods in other categories are validated under non-IID and partial client participation FL settings. The datasets used by all the methods can be listed as follows: LIBSVM, MNIST, FMNIST, FEMNIST, CIFAR-10, CIFAR-100, Tiny ImageNet, CINIC-10, SVHN, MNIST-8M, and others. These methods typically employ models such as logistic regression, MLP, DNN, or CNN, with objective functions like hinge loss or categorical cross-entropy loss.

Most methods across all categories are compared with both first-order and second-order FL methods. Based on the findings from their results, we observe that second-order methods generally improve FL training by more rapidly reducing classification error, training loss, test loss, and increasing test accuracy. This is achieved with fewer communication rounds or less time compared to first-order methods. From the results of FedDANE, it is noticed that DANE outperforms first-order FL methods only when the data is IID and when full client participation is used. Otherwise, it tends to underperform first-order methods such as FedAvg and FedProx, particularly when non-IID data is involved or with partial client participation.

So, the use of second-order information is highly beneficial for improving FL training compared to first-order methods, especially when aiming for a certain performance level from the global model with less time and fewer communication rounds. However, efficiently incorporating Hessian curvature remains a challenging task. This needs to be addressed in a way that minimizes computation and memory requirements, reduces transmission overhead, and is applicable to more realistic FL settings, such as non-convex objectives, non-IID data, and partial client participation.
%%%%%%%%%%%%%%%%%%%%%%%%%%%%%%%%%%%%%%%%%%%%%%%%%%%%%%
%%%%%%%%%%%%%%%%%%%%%%%%%%%%%%%%%%%%%%%%%%%%%%%%%%%%%%
%%%%%%%%%%%%%%%%%%%%%%%%%%%%%%%%%%%%%%%%%%%%%%%%%%%%%%
%%%%%%%%%%%%%%%%%%%%%%%%%%%%%%%%%%%%%%%%%%%%%%%%%%%%%% END of algorithmic discussion
\begin{table*}
    \centering
    \caption{Summary of all the methods under different categories}
    \renewcommand{\arraystretch}{1.0} % Adjust row spacing
    \begin{tabular}{|p{1.5cm}|p{15.5cm}|}
        \hline
        \multicolumn{1}{|c|}{\textbf{Category}} & \multicolumn{1}{c|}{\textbf{Summary}} \\ \hline  
        \centering\textbf{Hessian-free approach} 
        & These methods offer better convergence than first-order methods like FedAvg at similar computational costs but are limited to strongly convex and twice-differentiable objectives. These methods require multiple transmissions per round, increasing communication overhead, and are not suitable for Non-IID or partial client scenarios. They also demand higher local memory than FedAvg and struggle with limited data. Despite these challenges, they are effective in environments with well-distributed data and convex objectives \\ \hline
        \centering\textbf{Full Local Hessian Computation} 
        & These methods typically improve convergence rates, particularly with Non-IID data, and achieve super-linear convergence. However, they come with the trade-off of increased local and server computation and memory costs. While they reduce transmission costs compared to traditional Hessian transmission, they are still resource-intensive and may not guarantee convergence in all cases. These methods are generally suited for strongly convex and twice-differentiable objectives but can be less effective when dealing with limited data or highly variable client resources \\ \hline
        \centering\textbf{Quasi-Newton based} 
        & These methods typically offer better convergence, even for both convex and non-convex objectives with Non-IID data. However, they come with trade-offs, such as increased computation and memory costs on both the server and local clients. Additionally, they may experience instability during the initial stages of training if the model is far from the optimum. While these methods offer advantages in terms of faster convergence, particularly in non-convex settings, they require careful consideration of resource usage and initialization stability \\ \hline
        \centering\textbf{Hessian diagonal approximation} 
        & Methods under the Hessian Diagonal Approximation category aim to improve Federated Learning by approximating the Hessian matrix using its diagonal, reducing the need for storing and transmitting the full Hessian. These methods work well with both convex and non-convex objectives, particularly in non-IID and partial client scenarios. They offer comparable computation costs to first-order methods like FedAvg but slightly increase local and server memory costs and client-server transmission costs. However, they lack theoretical convergence proofs and ignore the off-diagonal elements of the Hessian, which contain valuable information about curvature. Despite these limitations, these methods are practical in resource-constrained settings but may not achieve optimal convergence. \\ \hline
        \centering\textbf{Nyström method-based Hessian approximation} 
        & These methods, such as FONN and Nys-FL, are efficient in handling non-IID and partial client scenarios and work well for both convex and non-convex objectives. They offer similar local and server computation costs to FedAvg, and both achieve linear-quadratic convergence. However, they have some trade-offs: FONN increases local memory costs and requires multiple transmissions per communication round, while Nys-FL raises server memory costs and may raise concerns about local data privacy. Both methods are limited to twice-differentiable objectives but provide practical solutions for large-scale federated learning environments. \\ \hline
        \centering\textbf{One-rank approximation of Hessian} 
        & Methods like FReNG and FAGH offer similar local and server computation costs to FedAvg and perform well in non-IID and partial client scenarios, avoiding full Hessian storage and transmission. Both methods achieve linear-quadratic convergence for convex objectives. However, FReNG is limited to probabilistic objectives and requires slightly higher server memory, while FAGH is restricted to twice-differentiable and non-linear objectives, increases both local and server memory costs, and raises client-to-server transmission costs. Despite these trade-offs, both methods provide efficient alternatives for large-scale federated learning in resource-constrained settings. \\ \hline
    \end{tabular}
    \label{tab:my_label4}
\end{table*}

\section{Conclusion}
Communication overhead resulting from the slow training of the global model is a significant challenge in federated learning (FL), which has not been adequately addressed in previous surveys of FL methods. In this paper, we investigate and categorize various second-order FL methods designed to accelerate FL training by utilizing Hessian curvature during the global model update. We describe how these methods leverage Hessian information while addressing its high complexities, and evaluate their advantages, disadvantages, applicability, effectiveness, and robustness across different FL settings. Finally, we provide both theoretical and empirical comparisons of these methods, highlighting their performance, efficiency, and potential for improving FL in real-world scenarios.   

\bibliographystyle{IEEEtran}
{\small
\bibliography{ref}}

\begin{thebibliography}{100}
\providecommand{\url}[1]{#1}
\csname url@samestyle\endcsname
\providecommand{\newblock}{\relax}
\providecommand{\bibinfo}[2]{#2}
\providecommand{\BIBentrySTDinterwordspacing}{\spaceskip=0pt\relax}
\providecommand{\BIBentryALTinterwordstretchfactor}{4}
\providecommand{\BIBentryALTinterwordspacing}{\spaceskip=\fontdimen2\font plus
\BIBentryALTinterwordstretchfactor\fontdimen3\font minus \fontdimen4\font\relax}
\providecommand{\BIBforeignlanguage}[2]{{%
\expandafter\ifx\csname l@#1\endcsname\relax
\typeout{** WARNING: IEEEtran.bst: No hyphenation pattern has been}%
\typeout{** loaded for the language `#1'. Using the pattern for}%
\typeout{** the default language instead.}%
\else
\language=\csname l@#1\endcsname
\fi
#2}}
\providecommand{\BIBdecl}{\relax}
\BIBdecl

\bibitem{mcmahan2016federated}
H.~B. McMahan, E.~Moore, D.~Ramage, and B.~A. y~Arcas, ``Federated learning of deep networks using model averaging,'' \emph{arXiv preprint arXiv:1602.05629}, vol.~2, no.~2, 2016.

\bibitem{MathewAfedprox_24}
C.~Mathew and P.~Asha, ``Fedprox: Fedsplit algorithm based federated learning for statistical and system heterogeneity in medical data communication,'' \emph{J. Internet Serv. Inf. Secur.}, vol.~14, no.~3, pp. 353--370, 2024.

\bibitem{KarimireddySCAFFOLD_20}
S.~P. Karimireddy, S.~Kale, M.~Mohri, S.~J. Reddi, S.~U. Stich, and A.~T. Suresh, ``{SCAFFOLD:} stochastic controlled averaging for federated learning,'' in \emph{Proceedings of the 37th International Conference on Machine Learning, {ICML} 2020, 13-18 July 2020, Virtual Event}, ser. Proceedings of Machine Learning Research, vol. 119.\hskip 1em plus 0.5em minus 0.4em\relax {PMLR}, 2020, pp. 5132--5143.

\bibitem{li2021_MOON}
Q.~Li, B.~He, and D.~Song, ``Model-contrastive federated learning,'' in \emph{Proceedings of the IEEE/CVF conference on computer vision and pattern recognition}, 2021, pp. 10\,713--10\,722.

\bibitem{AcarZNMWS21feddyn}
D.~A.~E. Acar, Y.~Zhao, R.~M. Navarro, M.~Mattina, P.~N. Whatmough, and V.~Saligrama, ``Federated learning based on dynamic regularization,'' in \emph{9th International Conference on Learning Representations, {ICLR} 2021, Virtual Event, Austria, May 3-7, 2021}.\hskip 1em plus 0.5em minus 0.4em\relax OpenReview.net, 2021.

\bibitem{barba2021implicitFedGA}
L.~Barba, M.~Jaggi, and Y.~Dandi, ``Implicit gradient alignment in distributed and federated learning,'' in \emph{AAAI Conference on Artificial Intelligence, AAAI}, vol.~22, 2021.

\bibitem{nagaraju2023handling}
C.~Nagaraju, M.~Sen, C.~K. Mohan, F.~Imai, C.~Distante, and S.~Battiato, ``Handling data heterogeneity in federated learning with global data distribution.'' in \emph{IMPROVE}, 2023, pp. 121--125.

\bibitem{yang2023littleSAFL}
H.~Yang, P.~Qiu, P.~Khanduri, and J.~Liu, ``With a little help from my friend: Server-aided federated learning with partial client participation,'' 2023.

\bibitem{yang2021achievinglienar_speed}
H.~Yang, M.~Fang, and J.~Liu, ``Achieving linear speedup with partial worker participation in non-iid federated learning,'' \emph{Proceedings of ICLR}, 2021.

\bibitem{gu2021fast}
X.~Gu, K.~Huang, J.~Zhang, and L.~Huang, ``Fast federated learning in the presence of arbitrary device unavailability,'' \emph{Advances in Neural Information Processing Systems}, vol.~34, pp. 12\,052--12\,064, 2021.

\bibitem{wu2023anchor}
F.~Wu, S.~Guo, Z.~Qu, S.~He, Z.~Liu, and J.~Gao, ``Anchor sampling for federated learning with partial client participation,'' in \emph{International Conference on Machine Learning}.\hskip 1em plus 0.5em minus 0.4em\relax PMLR, 2023, pp. 37\,379--37\,416.

\bibitem{jhunjhunwala2022fedvarp}
D.~Jhunjhunwala, P.~Sharma, A.~Nagarkatti, and G.~Joshi, ``Fedvarp: Tackling the variance due to partial client participation in federated learning,'' in \emph{Uncertainty in Artificial Intelligence}.\hskip 1em plus 0.5em minus 0.4em\relax PMLR, 2022, pp. 906--916.

\bibitem{wei2020federated_diff1}
K.~Wei, J.~Li, M.~Ding, C.~Ma, H.~H. Yang, F.~Farokhi, S.~Jin, T.~Q. Quek, and H.~V. Poor, ``Federated learning with differential privacy: Algorithms and performance analysis,'' \emph{IEEE transactions on information forensics and security}, vol.~15, pp. 3454--3469, 2020.

\bibitem{truex2020ldp}
S.~Truex, L.~Liu, K.-H. Chow, M.~E. Gursoy, and W.~Wei, ``Ldp-fed: Federated learning with local differential privacy,'' in \emph{Proceedings of the third ACM international workshop on edge systems, analytics and networking}, 2020, pp. 61--66.

\bibitem{adnan2022federated}
M.~Adnan, S.~Kalra, J.~C. Cresswell, G.~W. Taylor, and H.~R. Tizhoosh, ``Federated learning and differential privacy for medical image analysis,'' \emph{Scientific reports}, vol.~12, no.~1, p. 1953, 2022.

\bibitem{triastcyn2019federated}
A.~Triastcyn and B.~Faltings, ``Federated learning with bayesian differential privacy,'' in \emph{2019 IEEE International Conference on Big Data (Big Data)}.\hskip 1em plus 0.5em minus 0.4em\relax IEEE, 2019, pp. 2587--2596.

\bibitem{mohammadi2021differential}
N.~Mohammadi, J.~Bai, Q.~Fan, Y.~Song, Y.~Yi, and L.~Liu, ``Differential privacy meets federated learning under communication constraints,'' \emph{IEEE Internet of Things Journal}, vol.~9, no.~22, pp. 22\,204--22\,219, 2021.

\bibitem{wang2020attack}
H.~Wang, K.~Sreenivasan, S.~Rajput, H.~Vishwakarma, S.~Agarwal, J.-y. Sohn, K.~Lee, and D.~Papailiopoulos, ``Attack of the tails: Yes, you really can backdoor federated learning,'' \emph{Advances in neural information processing systems}, vol.~33, pp. 16\,070--16\,084, 2020.

\bibitem{darzi2024exploring}
E.~Darzi, F.~Dubost, N.~M. Sijtsema, and P.~M. van Ooijen, ``Exploring adversarial attacks in federated learning for medical imaging,'' \emph{IEEE Transactions on Industrial Informatics}, 2024.

\bibitem{song2020analyzing}
M.~Song, Z.~Wang, Z.~Zhang, Y.~Song, Q.~Wang, J.~Ren, and H.~Qi, ``Analyzing user-level privacy attack against federated learning,'' \emph{IEEE Journal on Selected Areas in Communications}, vol.~38, no.~10, pp. 2430--2444, 2020.

\bibitem{hu2023toward}
F.~Hu, W.~Zhou, K.~Liao, H.~Li, and D.~Tong, ``Toward federated learning models resistant to adversarial attacks,'' \emph{IEEE Internet of Things Journal}, vol.~10, no.~19, pp. 16\,917--16\,930, 2023.

\bibitem{chen2020zero}
Z.~Chen, P.~Tian, W.~Liao, and W.~Yu, ``Zero knowledge clustering based adversarial mitigation in heterogeneous federated learning,'' \emph{IEEE Transactions on Network Science and Engineering}, vol.~8, no.~2, pp. 1070--1083, 2020.

\bibitem{WangNeurIPSgiant_18}
S.~Wang, F.~Roosta, P.~Xu, and M.~W. Mahoney, ``Giant: Globally improved approximate newton method for distributed optimization,'' \emph{Advances in Neural Information Processing Systems}, vol.~31, 2018.

\bibitem{guptalocalnewton_21}
V.~Gupta, A.~Ghosh, M.~Derezinski, R.~Khanna, K.~Ramchandran, and M.~Mahoney, ``Localnewton: Reducing communication bottleneck for distributed learning,'' \emph{arXiv preprint arXiv:2105.07320}, 2021.

\bibitem{Mafedsso_22}
\BIBentryALTinterwordspacing
X.~Ma, R.~Bao, J.~Jiang, Y.~Liu, A.~Jiang, J.~Yan, X.~Liu, and Z.~Pan, ``Fedsso: A federated server-side second-order optimization algorithm,'' 2022. [Online]. Available: \url{https://arxiv.org/abs/2206.09576}
\BIBentrySTDinterwordspacing

\bibitem{liDANE_19}
O.~Shamir, N.~Srebro, and T.~Zhang, ``Communication-efficient distributed optimization using an approximate newton-type method,'' in \emph{Proceedings of the 31th International Conference on Machine Learning, {ICML} 2014, Beijing, China, 21-26 June 2014}, ser. {JMLR} Workshop and Conference Proceedings, vol.~32.\hskip 1em plus 0.5em minus 0.4em\relax JMLR.org, 2014, pp. 1000--1008.

\bibitem{safaryanfednl_22}
M.~Safaryan, R.~Islamov, X.~Qian, and P.~Richt{\'{a}}rik, ``Fednl: Making newton-type methods applicable to federated learning,'' in \emph{International Conference on Machine Learning, {ICML} 2022, 17-23 July 2022, Baltimore, Maryland, {USA}}, ser. Proceedings of Machine Learning Research, K.~Chaudhuri, S.~Jegelka, L.~Song, C.~Szepesv{\'{a}}ri, G.~Niu, and S.~Sabato, Eds., vol. 162.\hskip 1em plus 0.5em minus 0.4em\relax {PMLR}, 2022, pp. 18\,959--19\,010.

\bibitem{mrinmaynys-fl_23}
M.~Sen, C.~K. Mohan, and K.~Qin, ``Nys-fl: A communication efficient federated learning with nyström approximated global newton direction,'' in \emph{2023 IEEE International Conference on High Performance Computing \& Communications, Data Science \& Systems, Smart City \& Dependability in Sensor, Cloud \& Big Data Systems \& Application (HPCC/DSS/SmartCity/DependSys)}, 2023.

\bibitem{nagarajufonn_23}
C.~Nagaraju, M.~Sen, and C.~K. Mohan, ``Fonn: Federated optimization with nys-newton,'' in \emph{TENCON 2023-2023 IEEE Region 10 Conference (TENCON)}.\hskip 1em plus 0.5em minus 0.4em\relax IEEE, 2023, pp. 530--534.

\bibitem{yangover_the_air_22}
P.~Yang, Y.~Jiang, T.~Wang, Y.~Zhou, Y.~Shi, and C.~N. Jones, ``Over-the-air federated learning via second-order optimization,'' \emph{IEEE transactions on wireless communications}, vol.~21, no.~12, pp. 10\,560--10\,575, 2022.

\bibitem{senfoplahd_23}
M.~Sen and C.~K. Mohan, ``Foplahd: Federated optimization using locally approximated hessian diagonal,'' in \emph{International Conference on Big Data Analytics}.\hskip 1em plus 0.5em minus 0.4em\relax Springer, 2023, pp. 235--245.

\bibitem{dinhdone_22}
C.~T. Dinh, N.~H. Tran, T.~D. Nguyen, W.~Bao, A.~R. Balef, B.~B. Zhou, and A.~Y. Zomaya, ``Done: distributed approximate newton-type method for federated edge learning,'' \emph{IEEE Transactions on Parallel and Distributed Systems}, vol.~33, no.~11, pp. 2648--2660, 2022.

\bibitem{senfreng_23}
M.~Sen and C.~Gayatri, ``Freng: Federated optimization by using regularized natural gradient descent,'' in \emph{2023 International Conference on Machine Learning and Applications (ICMLA)}.\hskip 1em plus 0.5em minus 0.4em\relax IEEE, 2023, pp. 1889--1894.

\bibitem{Banabilahsurvey2_22}
S.~Banabilah, M.~Aloqaily, E.~Alsayed, N.~Malik, and Y.~Jararweh, ``Federated learning review: Fundamentals, enabling technologies, and future applications,'' \emph{Inf. Process. Manag.}, vol.~59, no.~6, p. 103061, 2022.

\bibitem{kumar2023impact}
K.~N. Kumar, C.~K. Mohan, and L.~R. Cenkeramaddi, ``The impact of adversarial attacks on federated learning: A survey,'' \emph{IEEE Transactions on Pattern Analysis and Machine Intelligence}, vol.~46, no.~5, pp. 2672--2691, 2023.

\bibitem{zhang2023systematic}
Y.~Zhang, Y.~Lu, and F.~Liu, ``A systematic survey for differential privacy techniques in federated learning,'' \emph{Journal of Information Security}, vol.~14, no.~2, pp. 111--135, 2023.

\bibitem{shan2024differential}
F.~Shan, S.~Mao, Y.~Lu, and S.~Li, ``Differential privacy federated learning: A comprehensive review.'' \emph{International Journal of Advanced Computer Science \& Applications}, vol.~15, no.~7, 2024.

\bibitem{fu2024differentially}
J.~Fu, Y.~Hong, X.~Ling, L.~Wang, X.~Ran, Z.~Sun, W.~H. Wang, Z.~Chen, and Y.~Cao, ``Differentially private federated learning: A systematic review,'' \emph{arXiv preprint arXiv:2405.08299}, 2024.

\bibitem{feng2025survey}
Y.~Feng, Y.~Guo, Y.~Hou, Y.~Wu, M.~Lao, T.~Yu, and G.~Liu, ``A survey of security threats in federated learning,'' \emph{Complex \& Intelligent Systems}, vol.~11, no.~2, pp. 1--26, 2025.

\bibitem{akhterchallengessurvey}
N.~AKHTER, M.~HASAN, R.~AMIN, and K.~E. AZIZ, ``Challenges, applications and design aspects of federated learning: A survey.''

\bibitem{zhusurvey4_21}
H.~Zhu, J.~Xu, S.~Liu, and Y.~Jin, ``Federated learning on non-iid data: A survey,'' \emph{Neurocomputing}, vol. 465, pp. 371--390, 2021.

\bibitem{lufederatedsurvey_24}
Z.~Lu, H.~Pan, Y.~Dai, X.~Si, and Y.~Zhang, ``Federated learning with non-iid data: A survey,'' \emph{IEEE Internet of Things Journal}, 2024.

\bibitem{hansurvey6_22}
X.~Han, M.~Gao, L.~Wang, Z.~He, and Y.~Wang, ``A survey of federated learning on non-iid data,'' \emph{ZTE Communications}, vol.~20, no.~3, p.~17, 2022.

\bibitem{HuangYSWLDY24FL_survey}
W.~Huang, M.~Ye, Z.~Shi, G.~Wan, H.~Li, B.~Du, and Q.~Yang, ``Federated learning for generalization, robustness, fairness: A survey and benchmark,'' \emph{{IEEE} Trans. Pattern Anal. Mach. Intell.}, vol.~46, no.~12, pp. 9387--9406, 2024.

\bibitem{zhaofederatedsurvey_18}
Y.~Zhao, M.~Li, L.~Lai, N.~Suda, D.~Civin, and V.~Chandra, ``Federated learning with non-iid data,'' \emph{arXiv preprint arXiv:1806.00582}, 2018.

\bibitem{bischoffsurvey_21}
S.~Bischoff, S.~G{\"u}nnemann, M.~Jaggi, and S.~U. Stich, ``On second-order optimization methods for federated learning,'' \emph{arXiv preprint arXiv:2109.02388}, 2021.

\bibitem{tan2019review}
H.~H. Tan and K.~H. Lim, ``Review of second-order optimization techniques in artificial neural networks backpropagation,'' in \emph{IOP conference series: materials science and engineering}, vol. 495, no.~1.\hskip 1em plus 0.5em minus 0.4em\relax IOP Publishing, 2019, p. 012003.

\bibitem{kashyap2022survey}
R.~Kashyap, ``A survey of deep learning optimizers--first and second order methods,'' \emph{arXiv preprint arXiv:2211.15596}, 2022.

\bibitem{Ma_stateoftheartsurvey3_22}
X.~Ma, J.~Zhu, Z.~Lin, S.~Chen, and Y.~Qin, ``A state-of-the-art survey on solving non-iid data in federated learning,'' \emph{Future Gener. Comput. Syst.}, vol. 135, pp. 244--258, 2022.

\bibitem{ketkarSGD_17}
N.~Ketkar and N.~Ketkar, ``Stochastic gradient descent,'' \emph{Deep learning with Python: A hands-on introduction}, pp. 113--132, 2017.

\bibitem{singh2021nys}
D.~Singh, H.~Tankaria, and M.~Yamada, ``Nys-newton: Nystr$\backslash$" om-approximated curvature for stochastic optimization,'' \emph{arXiv preprint arXiv:2110.08577}, 2021.

\bibitem{agarwal2017second}
N.~Agarwal, B.~Bullins, and E.~Hazan, ``Second-order stochastic optimization for machine learning in linear time,'' \emph{Journal of Machine Learning Research}, vol.~18, no. 116, pp. 1--40, 2017.

\bibitem{DDerezinskiM19}
M.~Derezinski and M.~W. Mahoney, ``Distributed estimation of the inverse hessian by determinantal averaging,'' in \emph{Advances in Neural Information Processing Systems 32: Annual Conference on Neural Information Processing Systems 2019, NeurIPS 2019, December 8-14, 2019, Vancouver, BC, Canada}, H.~M. Wallach, H.~Larochelle, A.~Beygelzimer, F.~d'Alch{\'{e}}{-}Buc, E.~B. Fox, and R.~Garnett, Eds., 2019, pp. 11\,401--11\,411.

\bibitem{martens2010deep}
J.~Martens \emph{et~al.}, ``Deep learning via hessian-free optimization.'' in \emph{Icml}, vol.~27, 2010, pp. 735--742.

\bibitem{cho2015hessian}
M.~Cho, C.~Dhir, and J.~Lee, ``Hessian-free optimization for learning deep multidimensional recurrent neural networks,'' \emph{Advances in Neural Information Processing Systems}, vol.~28, 2015.

\bibitem{shamirDANE_2014}
O.~Shamir, N.~Srebro, and T.~Zhang, ``Communication-efficient distributed optimization using an approximate newton-type method,'' in \emph{International conference on machine learning}.\hskip 1em plus 0.5em minus 0.4em\relax PMLR, 2014, pp. 1000--1008.

\bibitem{zhang2015disco}
Y.~Zhang and X.~Lin, ``Disco: Distributed optimization for self-concordant empirical loss,'' in \emph{International conference on machine learning}.\hskip 1em plus 0.5em minus 0.4em\relax PMLR, 2015, pp. 362--370.

\bibitem{reddi2016aide}
S.~J. Reddi, J.~Kone{\v{c}}n{\`y}, P.~Richt{\'a}rik, B.~P{\'o}cz{\'o}s, and A.~Smola, ``Aide: Fast and communication efficient distributed optimization,'' \emph{arXiv preprint arXiv:1608.06879}, 2016.

\bibitem{li2019feddane}
T.~Li, A.~K. Sahu, M.~Zaheer, M.~Sanjabi, A.~Talwalkar, and V.~Smithy, ``Feddane: A federated newton-type method,'' in \emph{2019 53rd Asilomar Conference on Signals, Systems, and Computers}.\hskip 1em plus 0.5em minus 0.4em\relax IEEE, 2019, pp. 1227--1231.

\bibitem{shalev2013stochastic}
S.~Shalev-Shwartz and T.~Zhang, ``Stochastic dual coordinate ascent methods for regularized loss,'' \emph{The Journal of Machine Learning Research}, vol.~14, no.~1, pp. 567--599, 2013.

\bibitem{lecun1998gradientmnist}
Y.~LeCun, L.~Bottou, Y.~Bengio, and P.~Haffner, ``Gradient-based learning applied to document recognition,'' \emph{Proceedings of the IEEE}, vol.~86, no.~11, pp. 2278--2324, 1998.

\bibitem{zinkevich2010parallelized}
M.~Zinkevich, M.~Weimer, L.~Li, and A.~Smola, ``Parallelized stochastic gradient descent,'' \emph{Advances in neural information processing systems}, vol.~23, 2010.

\bibitem{boyd2011distributed}
S.~Boyd, N.~Parikh, E.~Chu, B.~Peleato, J.~Eckstein \emph{et~al.}, ``Distributed optimization and statistical learning via the alternating direction method of multipliers,'' \emph{Foundations and Trends{\textregistered} in Machine learning}, vol.~3, no.~1, pp. 1--122, 2011.

\bibitem{qu2015quartz}
Z.~Qu, P.~Richt{\'a}rik, and T.~Zhang, ``Quartz: Randomized dual coordinate ascent with arbitrary sampling,'' \emph{Advances in neural information processing systems}, vol.~28, 2015.

\bibitem{lin2015universal}
H.~Lin, J.~Mairal, and Z.~Harchaoui, ``A universal catalyst for first-order optimization,'' \emph{Advances in neural information processing systems}, vol.~28, 2015.

\bibitem{chang2011libsvm}
C.-C. Chang and C.-J. Lin, ``Libsvm: a library for support vector machines,'' \emph{ACM transactions on intelligent systems and technology (TIST)}, vol.~2, no.~3, pp. 1--27, 2011.

\bibitem{takac2016distributed}
M.~Tak{\'a}c, ``Distributed optimization with arbitrary local solvers: Cocoa+ and beyond,'' 2016.

\bibitem{caldas2018leaf}
S.~Caldas, S.~M.~K. Duddu, P.~Wu, T.~Li, J.~Kone{\v{c}}n{\`y}, H.~B. McMahan, V.~Smith, and A.~Talwalkar, ``Leaf: A benchmark for federated settings,'' \emph{arXiv preprint arXiv:1812.01097}, 2018.

\bibitem{li2020federatedprox}
T.~Li, A.~K. Sahu, M.~Zaheer, M.~Sanjabi, A.~Talwalkar, and V.~Smith, ``Federated optimization in heterogeneous networks,'' \emph{Proceedings of Machine learning and systems}, vol.~2, pp. 429--450, 2020.

\bibitem{golub2013matrixPCG}
G.~H. Golub and C.~F. Van~Loan, \emph{Matrix computations}.\hskip 1em plus 0.5em minus 0.4em\relax JHU press, 2013.

\bibitem{nesterov2013introductory}
Y.~Nesterov, \emph{Introductory lectures on convex optimization: A basic course}.\hskip 1em plus 0.5em minus 0.4em\relax Springer Science \& Business Media, 2013, vol.~87.

\bibitem{liu1989limited}
D.~C. Liu and J.~Nocedal, ``On the limited memory bfgs method for large scale optimization,'' \emph{Mathematical programming}, vol.~45, no.~1, pp. 503--528, 1989.

\bibitem{nazareth2009conjugate}
J.~L. Nazareth, ``Conjugate gradient method,'' \emph{Wiley Interdisciplinary Reviews: Computational Statistics}, vol.~1, no.~3, pp. 348--353, 2009.

\bibitem{rheinboldt2009classical}
W.~C. Rheinboldt, ``Classical iterative methods for linear systems,'' \emph{Tech. Univ. Munich}, 2009.

\bibitem{cohen2017emnist}
G.~Cohen, S.~Afshar, J.~Tapson, and A.~Van~Schaik, ``Emnist: Extending mnist to handwritten letters,'' in \emph{2017 international joint conference on neural networks (IJCNN)}.\hskip 1em plus 0.5em minus 0.4em\relax IEEE, 2017, pp. 2921--2926.

\bibitem{anguita2013publicHAR}
D.~Anguita, A.~Ghio, L.~Oneto, X.~Parra, J.~L. Reyes-Ortiz \emph{et~al.}, ``A public domain dataset for human activity recognition using smartphones.'' in \emph{Esann}, vol.~3, no.~1, 2013, pp. 3--4.

\bibitem{qianbasis_matter_21}
X.~Qian, R.~Islamov, M.~Safaryan, and P.~Richt{\'{a}}rik, ``Basis matters: Better communication-efficient second order methods for federated learning,'' in \emph{International Conference on Artificial Intelligence and Statistics, {AISTATS} 2022, 28-30 March 2022, Virtual Event}, ser. Proceedings of Machine Learning Research, G.~Camps{-}Valls, F.~J.~R. Ruiz, and I.~Valera, Eds., vol. 151.\hskip 1em plus 0.5em minus 0.4em\relax {PMLR}, 2022, pp. 680--720.

\bibitem{elgabli_fednew_22}
A.~Elgabli, C.~B. Issaid, A.~S. Bedi, K.~Rajawat, M.~Bennis, and V.~Aggarwal, ``Fednew: A communication-efficient and privacy-preserving newton-type method for federated learning,'' in \emph{International conference on machine learning}.\hskip 1em plus 0.5em minus 0.4em\relax PMLR, 2022, pp. 5861--5877.

\bibitem{dal2024shed}
N.~Dal~Fabbro, S.~Dey, M.~Rossi, and L.~Schenato, ``Shed: A newton-type algorithm for federated learning based on incremental hessian eigenvector sharing,'' \emph{Automatica}, vol. 160, p. 111460, 2024.

\bibitem{li2024fedns}
J.~Li, Y.~Liu, and W.~Wang, ``Fedns: A fast sketching newton-type algorithm for federated learning,'' in \emph{Proceedings of the AAAI Conference on Artificial Intelligence}, vol.~38, no.~12, 2024, pp. 13\,509--13\,517.

\bibitem{NL1}
R.~Islamov, X.~Qian, and P.~Richt{\'a}rik, ``Distributed second order methods with fast rates and compressed communication,'' in \emph{International conference on machine learning}.\hskip 1em plus 0.5em minus 0.4em\relax PMLR, 2021, pp. 4617--4628.

\bibitem{DIANA_2024}
K.~Mishchenko, E.~Gorbunov, M.~Tak{\'a}{\v{c}}, and P.~Richt{\'a}rik, ``Distributed learning with compressed gradient differences,'' \emph{Optimization Methods and Software}, pp. 1--16, 2024.

\bibitem{ADIANA_li_20}
Z.~Li, D.~Kovalev, X.~Qian, and P.~Richt{\'a}rik, ``Acceleration for compressed gradient descent in distributed and federated optimization,'' \emph{arXiv preprint arXiv:2002.11364}, 2020.

\bibitem{gorbunov2021local}
E.~Gorbunov, F.~Hanzely, and P.~Richt{\'a}rik, ``Local sgd: Unified theory and new efficient methods,'' in \emph{International Conference on Artificial Intelligence and Statistics}.\hskip 1em plus 0.5em minus 0.4em\relax PMLR, 2021, pp. 3556--3564.

\bibitem{crane2019dingo}
R.~Crane and F.~Roosta, ``Dingo: Distributed newton-type method for gradient-norm optimization,'' \emph{Advances in neural information processing systems}, vol.~32, 2019.

\bibitem{nocedal2006stephenn_numerical_optimization}
J.~Nocedal, ``Stephen j,'' \emph{Wright. Numerical optimization. Springer Science+ Business Media}, 2006.

\bibitem{hamidi2025distributed}
S.~M. Hamidi and L.~Ye, ``Distributed quasi-newton method for fair and fast federated learning,'' \emph{arXiv preprint arXiv:2501.10877}, 2025.

\bibitem{krizhevsky2009learning_cifar10}
A.~Krizhevsky, G.~Hinton \emph{et~al.}, ``Learning multiple layers of features from tiny images,'' 2009.

\bibitem{yuan2020federated_AC}
H.~Yuan and T.~Ma, ``Federated accelerated stochastic gradient descent,'' \emph{Advances in Neural Information Processing Systems}, vol.~33, pp. 5332--5344, 2020.

\bibitem{reddi2020adaptive_opt}
S.~Reddi, Z.~Charles, M.~Zaheer, Z.~Garrett, K.~Rush, J.~Kone{\v{c}}n{\`y}, S.~Kumar, and H.~B. McMahan, ``Adaptive federated optimization,'' \emph{arXiv preprint arXiv:2003.00295}, 2020.

\bibitem{lifeddane_19}
T.~Li, A.~K. Sahu, M.~Zaheer, M.~Sanjabi, A.~Talwalkar, and V.~Smithy, ``Feddane: A federated newton-type method,'' in \emph{2019 53rd Asilomar Conference on Signals, Systems, and Computers}.\hskip 1em plus 0.5em minus 0.4em\relax IEEE, 2019, pp. 1227--1231.

\bibitem{mukai2003algorithms}
H.~Mukai, ``Algorithms for multicriterion optimization,'' \emph{IEEE transactions on automatic control}, vol.~25, no.~2, pp. 177--186, 2003.

\bibitem{le2015tiny}
Y.~Le and X.~Yang, ``Tiny imagenet visual recognition challenge,'' \emph{CS 231N}, vol.~7, no.~7, p.~3, 2015.

\bibitem{darlow2018cinic}
L.~N. Darlow, E.~J. Crowley, A.~Antoniou, and A.~J. Storkey, ``Cinic-10 is not imagenet or cifar-10,'' \emph{arXiv preprint arXiv:1810.03505}, 2018.

\bibitem{wang2021federated}
Z.~Wang, X.~Fan, J.~Qi, C.~Wen, C.~Wang, and R.~Yu, ``Federated learning with fair averaging,'' \emph{arXiv preprint arXiv:2104.14937}, 2021.

\bibitem{wang2020federated}
H.~Wang, M.~Yurochkin, Y.~Sun, D.~Papailiopoulos, and Y.~Khazaeni, ``Federated learning with matched averaging,'' \emph{arXiv preprint arXiv:2002.06440}, 2020.

\bibitem{he2016deep}
K.~He, X.~Zhang, S.~Ren, and J.~Sun, ``Deep residual learning for image recognition,'' in \emph{Proceedings of the IEEE conference on computer vision and pattern recognition}, 2016, pp. 770--778.

\bibitem{senpreml_23}
M.~Sen, C.~K. Mohan, and A.~K. Qin, ``Federated optimization with linear-time approximated hessian diagonal,'' in \emph{International Conference on Pattern Recognition and Machine Intelligence}.\hskip 1em plus 0.5em minus 0.4em\relax Springer, 2023, pp. 106--113.

\bibitem{ahmad2023robust}
A.~Ahmad, W.~Luo, and A.~Robles-Kelly, ``Robust federated learning under statistical heterogeneity via hessian-weighted aggregation,'' \emph{Machine Learning}, vol. 112, no.~2, pp. 633--654, 2023.

\bibitem{elbakary2024fed}
A.~Elbakary, C.~B. Issaid, M.~Shehab, K.~Seddik, T.~ElBatt, and M.~Bennis, ``Fed-sophia: A communication-efficient second-order federated learning algorithm,'' in \emph{ICC 2024-IEEE International Conference on Communications}.\hskip 1em plus 0.5em minus 0.4em\relax IEEE, 2024, pp. 950--955.

\bibitem{bekas2007estimator}
C.~Bekas, E.~Kokiopoulou, and Y.~Saad, ``An estimator for the diagonal of a matrix,'' \emph{Applied numerical mathematics}, vol.~57, no. 11-12, pp. 1214--1229, 2007.

\bibitem{shoham2019overcoming}
N.~Shoham, T.~Avidor, A.~Keren, N.~Israel, D.~Benditkis, L.~Mor-Yosef, and I.~Zeitak, ``Overcoming forgetting in federated learning on non-iid data,'' \emph{arXiv preprint arXiv:1910.07796}, 2019.

\bibitem{liu2023sophia}
H.~Liu, Z.~Li, D.~Hall, P.~Liang, and T.~Ma, ``Sophia: A scalable stochastic second-order optimizer for language model pre-training,'' \emph{arXiv preprint arXiv:2305.14342}, 2023.

\bibitem{sen2024fagh}
M.~Sen \emph{et~al.}, ``Fagh: Accelerating federated learning with approximated global hessian,'' \emph{arXiv preprint arXiv:2403.11041}, 2024.

\bibitem{sun2023understanding}
Y.~Sun, L.~Shen, and D.~Tao, ``Understanding how consistency works in federated learning via stage-wise relaxed initialization,'' \emph{Advances in Neural Information Processing Systems}, vol.~36, pp. 80\,543--80\,574, 2023.

\bibitem{dandi2022implicit}
Y.~Dandi, L.~Barba, and M.~Jaggi, ``Implicit gradient alignment in distributed and federated learning,'' in \emph{Proceedings of the AAAI Conference on Artificial Intelligence}, vol.~36, no.~6, 2022, pp. 6454--6462.

\bibitem{jhunjhunwala2023fedexp}
D.~Jhunjhunwala, S.~Wang, and G.~Joshi, ``Fedexp: Speeding up federated averaging via extrapolation,'' in \emph{International Conference on Learning Representations}, 2023.

\end{thebibliography}

\begin{IEEEbiography}[{\includegraphics[width=1in,height=1.25in,clip,keepaspectratio]{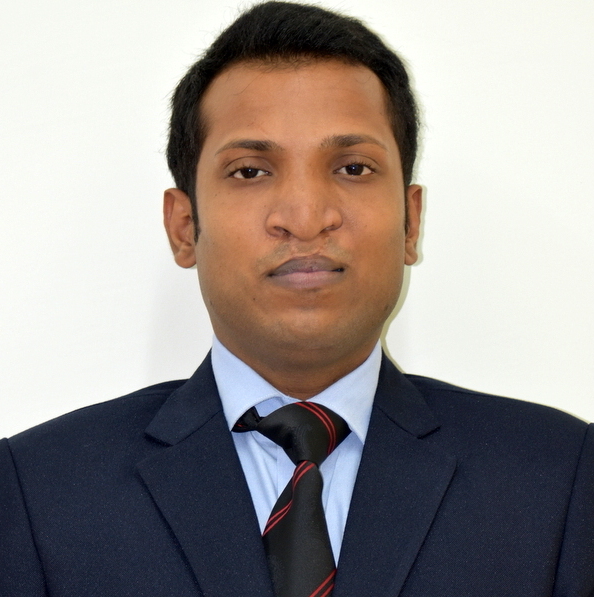}}]%
{Mrinmay Sen} is a joint research scholar at the Department of Artificial Intelligence, Indian Institute of Technology Hyderabad, India, and the Department of Computing Technologies, Swinburne University of Technology, Australia. He holds an M.Tech. degree from the Indian Institute of Technology Dhanbad, India, and a B.Eng. degree from Jadavpur University, Kolkata, India. His research interests focus on second-order optimization, federated learning and its practical applications, along with advancements in Computer Vision and deep learning. 
\end{IEEEbiography}

\begin{IEEEbiography}[{\includegraphics[width=1in,height=1.25in,clip,keepaspectratio]{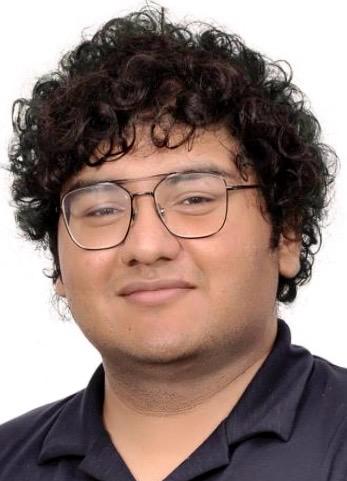}}]%
{Sidhant R Nair} is pursuing his 3rd year B.Tech. in Mechanical Engineering at the Indian Institute of Technology Delhi, India. His research interests focus on computer vision, second-order optimization, federated optimization and robotics.
\end{IEEEbiography}

\begin{IEEEbiography}[{\includegraphics[width=1in,height=1.25in,clip,keepaspectratio]{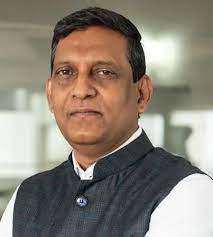}}]%
{C Krishna Mohan} is currently a Professor in the Department of Computer Science and Engineering at the Indian Institute of Technology Hyderabad, India. His research interests primarily include video content analysis, computer vision, machine learning, and deep learning. He has published over 90 papers in various international journals and conference proceedings. He is a Senior Member of IEEE, a Member of ACM, an AAAI Member, and a Life Member of ISTE.
\end{IEEEbiography}

\end{document}